\documentclass{article} 
\ifdefined\pdfminorversion\pdfminorversion=7\fi

\usepackage{iclr2027_conference,times}
\iclrfinalcopy
\date{}

\usepackage[utf8]{inputenc} 
\usepackage[T1]{fontenc}    
\usepackage{hyperref}       
\usepackage{url}            
\usepackage{graphicx}       
\usepackage{booktabs}       
\usepackage{amsfonts}       
\usepackage{nicefrac}       
\usepackage{microtype}      
\usepackage{xcolor}         
\usepackage{pifont}
\usepackage{amsmath}
\usepackage{amssymb}
\usepackage{mathtools}
\usepackage{amsthm}
\usepackage{multirow}
\usepackage{xspace}
\usepackage{enumitem}
\usepackage{placeins}
\usepackage[most]{tcolorbox}
\usepackage{courier}
\usepackage{wrapfig}
\usepackage[ruled,vlined]{algorithm2e}
\hypersetup{
    colorlinks=true,
    linkcolor=red,
    citecolor=cyan,
    filecolor=magenta,      
    urlcolor=magenta,
    }

\usepackage[font=small,skip=4pt]{caption}
\setlist{topsep=2pt, partopsep=0pt, itemsep=1pt, parsep=0pt, leftmargin=1.5em}
\renewcommand{\arraystretch}{0.9}
\makeatletter
\renewcommand{\paragraph}{\@startsection{paragraph}{4}{\z@}{0.6ex plus 0.3ex
  minus .2ex}{-1em}{\normalsize\bf}}
\let\backmattersection\section
\let\backmattersubsection\subsection
\let\backmatterparagraph\paragraph
\newcommand{\compactbodyspacing}{%
  \setlength{\parskip}{1.9pt plus .3pt minus .2pt}%
  \setlength{\abovedisplayskip}{3.6pt plus 1pt minus 1pt}%
  \setlength{\belowdisplayskip}{3.6pt plus 1pt minus 1pt}%
  \setlength{\belowdisplayshortskip}{2pt plus 1pt minus 1pt}%
  \setlength{\textfloatsep}{8pt plus 1pt minus 2pt}%
  \setlength{\floatsep}{8pt plus 1pt minus 1pt}%
  \setlength{\intextsep}{8pt plus 1pt minus 1pt}%
  \renewcommand{\section}{\@startsection{section}{1}{\z@}
    {-5pt plus -1pt minus -.5pt}{3.5pt plus .5pt}{\large\sc\raggedright}}%
  \renewcommand{\subsection}{\@startsection{subsection}{2}{\z@}
    {-4.5pt plus -.5pt minus -.5pt}{2pt plus .5pt}{\normalsize\sc\raggedright}}%
  \renewcommand{\paragraph}{\@startsection{paragraph}{4}{\z@}
    {.2pt plus .1pt minus .1pt}{-1em}{\normalsize\bf}}%
}
\newcommand{\restorebackmatterspacing}{%
  \let\section\backmattersection
  \let\subsection\backmattersubsection
  \let\paragraph\backmatterparagraph
  \setlength{\parskip}{4.5pt plus 1pt minus 1pt}%
  \setlength{\abovedisplayskip}{4pt plus 2pt minus 2pt}%
  \setlength{\belowdisplayskip}{4pt plus 2pt minus 2pt}%
  \setlength{\belowdisplayshortskip}{3pt plus 2pt minus 2pt}%
  \setlength{\textfloatsep}{10pt plus 2pt minus 3pt}%
  \setlength{\floatsep}{10pt plus 2pt minus 2pt}%
  \setlength{\intextsep}{10pt plus 2pt minus 2pt}%
}
\makeatother
\newcommand{\twfive}{\dimexpr\linewidth-24pt\relax}  
\newcommand{\twsix}{\dimexpr\linewidth-30pt\relax}   
\usepackage{array}
\newcolumntype{L}[1]{>{\raggedright\arraybackslash}p{#1}}

\SetKwComment{Comment}{$\triangleright$\ }{}
\newcommand{\llmname}[1]{{\fontfamily{pcr}\selectfont {#1}}\xspace}
\newcommand{\ourmethod}{{\fontfamily{lmtt}\selectfont \textbf{LatentWeave}}\xspace}

\title{LatentWeave: Test-Time Latent Optimization \\ for Multimodal Reasoning}

\author{%
\parbox[t]{\dimexpr\textwidth-2\tabcolsep\relax}{\raggedright
{\bfseries
\mbox{Fanci Meng\textsuperscript{1}}\quad
\mbox{Kun Wang\textsuperscript{2,$\dagger$}}\quad
\mbox{Guibin Zhang\textsuperscript{5}}\quad
\mbox{Xinlei Yu\textsuperscript{3}}\quad
\mbox{Jiaming Zhang\textsuperscript{2}}\quad
\mbox{Ming Li\textsuperscript{4}}\quad
\mbox{Yang Wang\textsuperscript{1,$\dagger$}}\endgraf}
\vspace{3pt}
{\normalfont
\mbox{\textsuperscript{1}University of Science and Technology of China}\quad
\mbox{\textsuperscript{2}Nanyang Technological University}\quad
\mbox{\textsuperscript{3}The Chinese University of Hong Kong}\quad
\mbox{\textsuperscript{4}The Chinese University of Hong Kong, Shenzhen}\quad
\mbox{\textsuperscript{5}National University of Singapore}\endgraf}
\vspace{3pt}
{\normalfont\texttt{fanc1m@mail.ustc.edu.cn}\endgraf}
}
}
\hypersetup{pdftitle={LatentWeave: Test-Time Latent Optimization for Multimodal Reasoning},
pdfauthor={Fanci Meng, Kun Wang, Guibin Zhang, Xinlei Yu, Jiaming Zhang, Ming Li, Yang Wang}}
\begin{document}

\maketitle
\begingroup
\renewcommand{\thefootnote}{\fnsymbol{footnote}}
\footnotetext[2]{Kun Wang and Yang Wang are the corresponding authors.}
\endgroup
\suppressfloats[t]

\begin{abstract}
  Recent advances in test-time optimization have shown that the multimodal reasoning performance of large multimodal language models (MLLMs) can be improved during inference without modifying backbone model parameters. However, existing approaches primarily operate at the prompt or output level, leaving the latent reasoning process largely uncontrolled and insufficient to prevent error accumulation during autoregressive generation. To address this limitation, we propose \ourmethod, a test-time framework that treats multimodal reasoning as a controllable latent trajectory. \ourmethod performs structured \textit{Latent Weaving}, progressively refining intermediate hidden states segment-wise to correct emerging deviations, and further introduces \textit{Reasoning Anchors} via lightweight WEAVE tokens to stabilize reasoning structure and guide answer-oriented generation. These mechanisms are designed to enable fine-grained, step-aware intervention in the latent reasoning process while keeping the main MLLM backbone parameters fixed. Extensive experiments across diverse multimodal reasoning benchmarks report improvements over the evaluated baselines; the contribution of latent optimization, anchor training, and additional test-time computation is analyzed separately in the ablation and efficiency studies. Our code is available at \url{https://github.com/jins7/LatentWeave}.
\end{abstract}

\section{Introduction}
Multimodal large language models (MLLMs) \citep{yin2024survey} increasingly support visual understanding and generation within unified systems \citep{ma2025unitok, wu2025janus}, alongside specialized capabilities such as image-quality reasoning \citep{li2025q}. These advances have broadened their use as general-purpose foundation models \citep{yang2023dawn, comanici2025gemini}. Research is consequently expanding from perception to multi-step reasoning over multimodal information \citep{li2025perception, viveiros2026lantern}. Maintaining a structured reasoning process remains challenging: LLaVA-CoT studies staged visual reasoning \citep{xu2025llava}, while CoMT evaluates the integration of visual operations into intermediate reasoning \citep{cheng2025comt}. As reasoning traces grow longer, the evolving evidence and intermediate representations become important targets for inference-time refinement \citep{jiang2026ttsp, zhang2026unsilencing}.

\begin{figure}[t]
    \centering
    \includegraphics[width=0.74\textwidth]{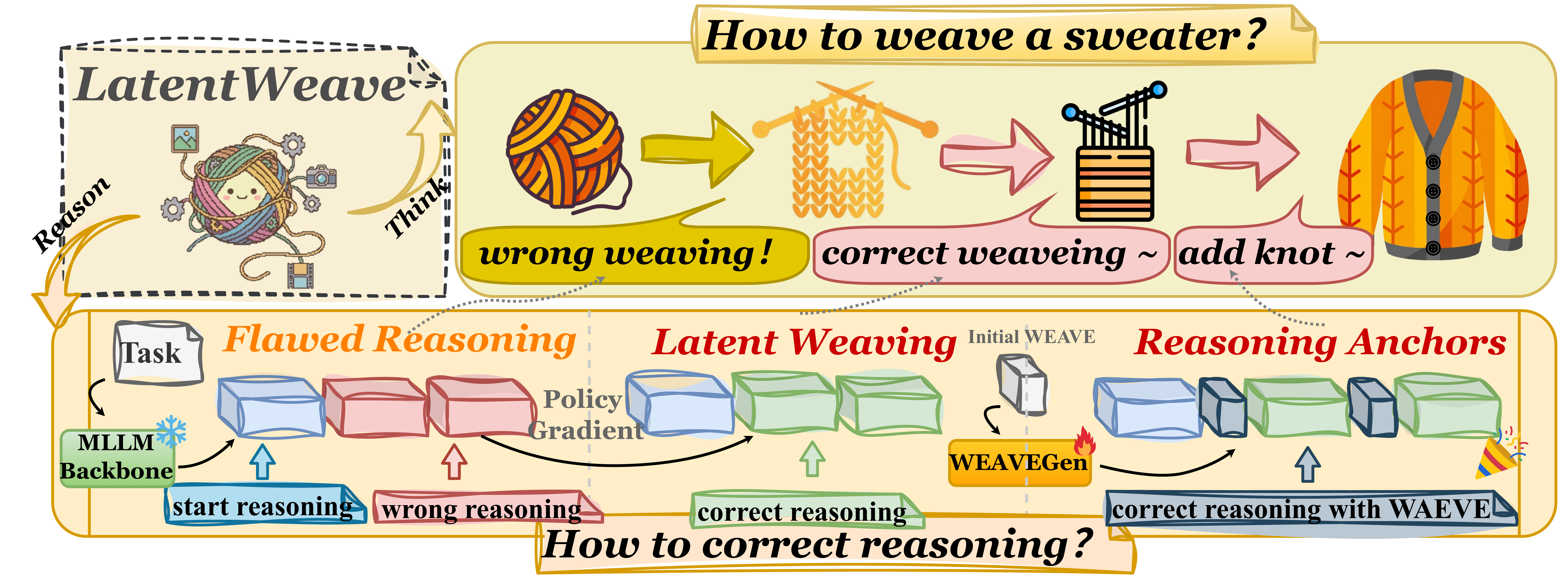}
    \caption{\ourmethod's Latent Weaving functions much like rectifying faulty knitting, preventing accumulation of errors and subsequent failure, whereas Reasoning Anchors mirror adding buttons at appropriate positions to a finished sweater body, enhancing its structural integrity.}
    \label{fig:intro}
\end{figure}

To improve the test-time performance of multimodal reasoning \citep{zhu2025internvl3, jiang2026ttsp}, existing studies have explored intervention strategies from three core dimensions: \ding{182} \textbf{Output-level refinement:} Focuses on answer post-processing and self-correction \citep{madaan2023selfrefineiterativerefinementselffeedback}, iteratively optimizing model-generated outputs via answer rewriting, candidate filtering, or reasoning-trace refinement \citep{jiang2026trajectory}. \ding{183} \textbf{Prompt-level control:} Leverages concise guidance from prompts or reasoning structures \citep{xia2026acts}, guiding models to follow predefined reasoning paths by constructing explicit reasoning chains \citep{wang2025multimodal}, decomposing sub-problems \citep{acuna2025socratic}, or introducing stepwise textual guidance \citep{xia2026acts}. \ding{184} \textbf{Parameter-level adaptation:} Relies on parameter updates and adaptive adjustments \citep{zhao2025attention}, incorporating reinforcement updates \citep{zuo2025ttrl} or temporary adapter optimization \citep{yuan2026relat} during inference to enhance performance on specific reasoning tasks. Although these strategies yield improvements, their primary intervention targets are outputs, prompts, or parameters rather than individual latent states. Recent methods also optimize continuous representations \citep{li2025seek, zhang2026unsilencing}, motivating step-aware intervention in multimodal reasoning.

These observations motivate treating multimodal reasoning as a control problem over latent trajectories. Output- and prompt-oriented methods address errors through output refinement or external guidance \citep{jiang2026trajectory, xia2026acts}. We hypothesize that some errors accumulate in intermediate hidden states and become harder to correct during subsequent generation. This motivates refining the evolving trajectory directly \citep{zhang2025latentevolve, zhang2026unsilencing}. Here, latent reasoning refers to iteratively refining the model's internal hidden states during autoregressive generation. Such intervention can revise intermediate decisions while retaining the pretrained backbone parameters. The central question is how to localize these updates, preserve useful preceding context, and propagate corrections into the remaining continuation.

Given the prior limitations, we propose \ourmethod, a test-time framework that models multimodal reasoning as an evolving latent trajectory and directly optimizes its hidden states. As in knitting a sweater (Figure~\ref{fig:intro}), an early deviation can be amplified by subsequent steps and distort the final result. Based on this insight, \ourmethod regulates reasoning trajectories via two paired core mechanisms: \textbf{Latent Weaving} refines hidden-state segments during generation. Unlike ACTS \citep{xia2026acts}, which steers a frozen reasoner through textual strategies and starter phrases, \ourmethod directly optimizes each active segment, preserves accepted segments, and re-decodes the remaining suffix. Rectified trajectories supervise the lightweight \textit{WEAVEGen} model to generate textual reasoning anchors. \textbf{Reasoning Anchors} differ from PILOT's latent-vector injection \citep{zheng2026pilot}: WEAVEGen serializes its guidance with \texttt{<WEAVE>} delimiters and passes it through the backbone's ordinary textual context, avoiding hidden-dimension transfer between models. WEAVEGen is trained offline and generates one anchor sequence per input. Both mechanisms keep the evaluated backbone fixed at test time; auxiliary generation costs are reported separately.

We summarize our contributions in the following three points:

\ding{182} \textbf{\textit{Problem formulation.}} We formulate multimodal reasoning as a latent trajectory control problem, hypothesizing that some reasoning failures arise from progressively accumulated deviations in intermediate latent states.

\ding{183} \textbf{\textit{Method.}} We propose \ourmethod, a frozen-backbone test-time framework that directly intervenes in latent representations during autoregressive generation. \textbf{Latent Weaving} refines hidden-state segments in sequence, while textual \textbf{Reasoning Anchors} are generated by a lightweight model (\textit{WEAVEGen}) trained on rectified trajectories. Their text-based interface supports backbones with different hidden dimensions.

\ding{184} \textbf{\textit{Experimental evaluation.}} Experiments cover seven benchmarks and five backbones (4B--27B). Over vanilla decoding, \ourmethod gains $6.57$ percentage points on MathVision with \llmname{Qwen3.5-4B}, and $6.57$ and $5.22$ points on MathVision and RealWorldQA, respectively, with \llmname{Qwen3.6-27B}, demonstrating improvements across model scales and task domains.

\begin{figure}[!t]
    \centering
    \includegraphics[width=0.88\linewidth]{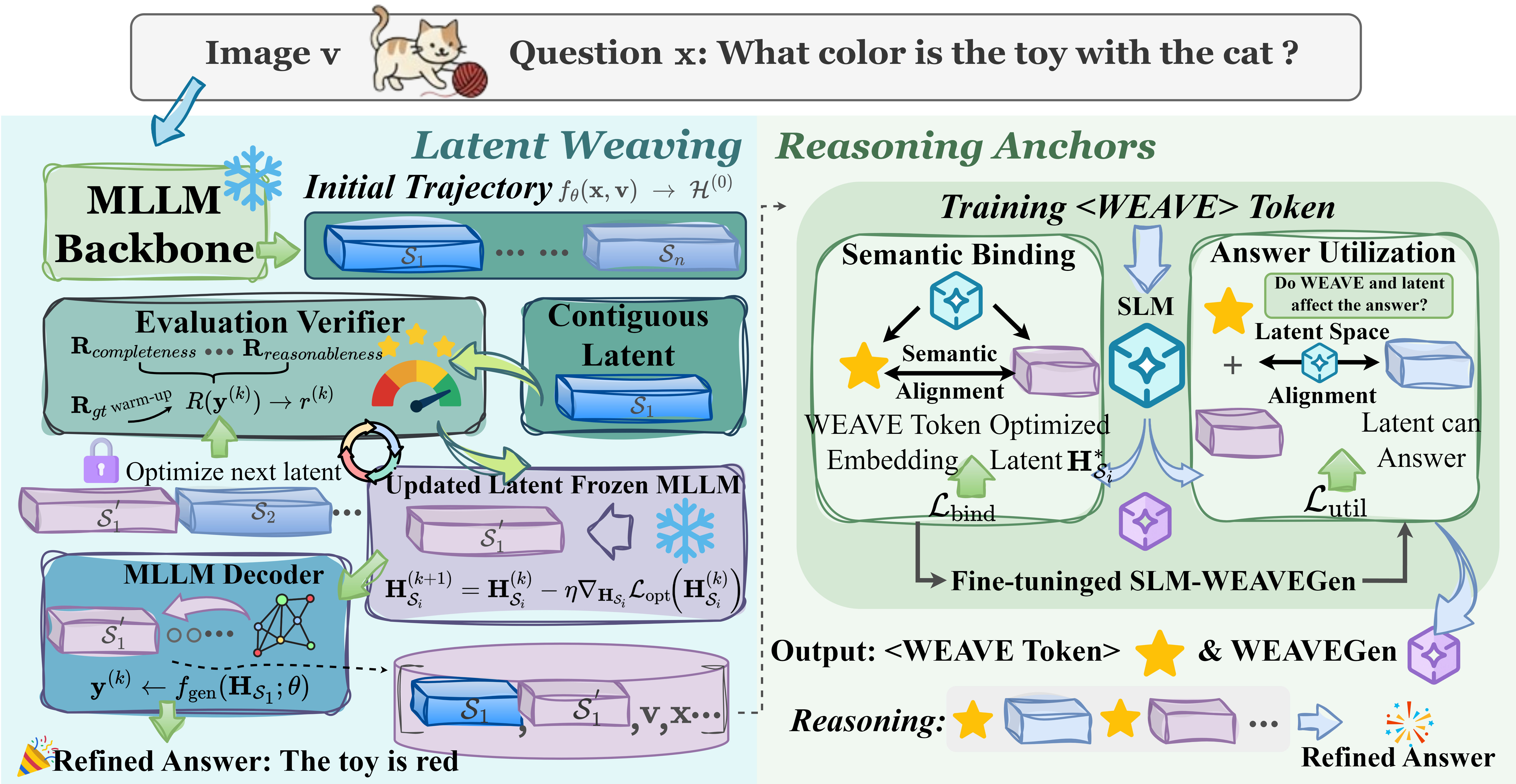}
    \caption{The overall framework of our proposed \ourmethod.}
    \label{fig:method}
\end{figure}

\section{Related Work}

\paragraph{Multimodal reasoning representations.} Visual Sketchpad uses executable sketches as intermediate visual reasoning steps \citep{hu2024visual}, while Coconut establishes continuous hidden-state reasoning in language models \citep{hao2024coconuttraininglargelanguagemodels}. Recent multimodal methods learn task-relevant latent representations: LanteRn interleaves text with visually grounded latent blocks \citep{viveiros2026lantern}; Pearl internalizes tool-use trajectories through predictive embedding objectives \citep{adhikari2026pearl}; and SLVR enriches region-level visual latents with semantic supervision and multi-query reinforcement learning \citep{xu2026slvr}. These methods principally learn how to generate useful representations. Complementing this line, \citet{li2026imagination} identify weak input-to-latent and latent-to-answer dependencies in the visual reasoners they examine. Their findings motivate checking whether intermediate states influence answers, rather than equating latent tokens with effective reasoning.

\paragraph{Test-time computation and adaptation.} Additional inference computation can target visual evidence, reasoning traces, or model parameters. TTSP explores visual-tool trajectories and maintains a correctable evidence ledger \citep{jiang2026ttsp}. Trajectory-guided structured sampling retrieves reasoning patterns and refines discrete traces through structured MCMC \citep{jiang2026trajectory}. These approaches improve explicit evidence or generated sequences. Parameter adaptation is a separate axis: TTRL updates model parameters using rewards estimated from unlabeled test data \citep{zuo2025ttrl}, while ReLAT adapts temporary LoRA parameters using a question-reconstruction objective for latent collaboration \citep{yuan2026relat}. Thus, test-time adaptation need not preserve model parameters. \ourmethod instead updates instance-specific hidden states while keeping the evaluated backbone and the offline-trained anchor generator fixed during evaluation.

\paragraph{Continuous test-time optimization.} LatentSeek applies policy-gradient updates to hidden states \citep{li2025seek}, and LatentEvolve incorporates retrieval and consolidation of prior latent optimizations \citep{zhang2025latentevolve}. More recently, $\nabla$-Reasoner optimizes continuous token logits under reward and likelihood objectives during iterative decoding \citep{wang2026nablareasoner}; its optimization variables are relaxed output representations. In multimodal models, \citet{zhang2026unsilencing} combine visual alignment with reward-guided optimization of reserved latent slots before answer decoding, keeping the backbone frozen. These are direct predecessors for inference-time continuous refinement. \ourmethod focuses on sequentially optimizing hidden-state segments of a multimodal generation trajectory, preserving accepted segments and regenerating the suffix after each update, with textual reasoning anchors guiding the generation of the remaining suffix.

\paragraph{Learned guidance and structural supervision.} PILOT trains an auxiliary network to produce a query-conditioned latent anchor for a frozen language model \citep{zheng2026pilot}. ACTS trains a controller that supplies strategies and starter phrases to a frozen reasoner at successive reasoning steps \citep{xia2026acts}. PMPS uses shared reasoning prototypes to align latent and explicit chains of thought during training \citep{li2026pmps}. These works establish precedents for latent anchoring, textual steering, and structural supervision. We combine segment-wise multimodal hidden-state optimization with an offline-trained textual guidance mechanism: WEAVEGen learns from optimized training trajectories, generates one textual anchor sequence per input before refinement, and conditions the evaluated backbone through ordinary text. This interface differs from PILOT's latent injection and ACTS's online stepwise controller, while online optimization refines the backbone's hidden states and keeps all auxiliary-network parameters fixed throughout evaluation.

\section{Methodology}
\label{sec:method}
The overall workflow of our method is illustrated in Figure~\ref{fig:method}. Without updating the evaluated backbone parameters online, \ourmethod operates directly in the hidden-state space to incrementally optimize reasoning-critical regions. This latent-level control allows the reasoning process to be adjusted at inference time while preserving the pretrained backbone. Section~\ref{pre} introduces the autoregressive formulation and latent optimization. Section~\ref{latent-weave} presents \textbf{\emph{Latent Weaving}}, which is used during offline trajectory collection and evaluation-time refinement. Section~\ref{reward} describes the reward signals guiding latent weaving, and Section~\ref{anchors} introduces \textbf{\emph{Reasoning Anchors}}, an offline-trained auxiliary generator whose textual output conditions the evaluated backbone. The overall procedure is summarized in Algorithm~\ref{alg:ourmethod} in Appendix~\ref{sec:appendix_algorithm}, which separates offline trajectory collection (Phase~I), offline anchor training (Phase~II), and evaluation-time latent refinement (Phase~III). Additional implementation details, notations, and experimental configurations are provided in Appendix~\ref{sec:appendix_full_method}.

\subsection{Preliminary}\label{pre}
We begin by introducing the multimodal autoregressive formulation and the latent representations optimized in subsequent sections, using vision--language backbones such as Qwen2.5-VL \citep{bai2025qwen2}. This formulation provides the foundation for the latent trajectory manipulation performed during offline collection and evaluation-time refinement in Algorithm~\ref{alg:ourmethod}. The notation is summarized in Table~\ref{tab:appendix_notation} in Appendix~\ref{sec:appendix_full_method}.

\paragraph{Latent Formulation.}
We study multimodal autoregressive generation given textual input $\mathbf{x}$ and visual context $\mathbf{v}$. Given a pretrained model $f_\theta$, decoding runs token by token, producing intermediate hidden states that encode the multimodal inputs and the generated prefix. In \ourmethod, these hidden states are viewed as latent variables and explicitly optimized during inference.

At decoding step $t$, the model computes a hidden state $\mathbf{h}_t$ from the input modalities and previously generated tokens, and predicts the next-token distribution as:
\begin{equation}
\begin{gathered}
\mathbf{h}_t = f_\theta(\mathbf{x}, \mathbf{v}, \mathbf{y}_{<t}), \\
p_\theta(\cdot \mid \mathbf{x}, \mathbf{v}, \mathbf{y}_{<t})
= \operatorname{softmax}(\mathbf{W}_o \mathbf{h}_t) ,
\end{gathered}
\label{eq:autoregressive}
\end{equation}
where $f_\theta$ denotes the pretrained vision--language model and $\mathbf{W}_o$ is the output projection matrix. Unlike standard decoding, \ourmethod exposes $\mathbf{h}_t$ as manipulable latent variables rather than treating them as fixed activations throughout generation.

The probability of an output sequence $\mathbf{y} = (y_1,\dots,y_T)$ factorizes autoregressively as: 
\begin{equation}
p_\theta(\mathbf{y} \mid \mathbf{x}, \mathbf{v})
= \prod_{t=1}^{T} p_\theta(y_t \mid \mathbf{x}, \mathbf{v}, \mathbf{y}_{<t}) .
\label{eq:sequence-prob}
\end{equation}

\paragraph{Test-Time Latent Optimization.}
Rather than updating model parameters at inference time, we directly operate on the latent hidden-state sequence $\mathbf{H} = (\mathbf{h}_1,\ldots,\mathbf{h}_T)$ defined in Eq.~\eqref{eq:autoregressive}. We test whether refining these states can correct reasoning errors with the backbone fixed. The objective is to find a latent trajectory maximizing the task-specific reward of its generated output:
\begin{equation}
\mathbf{H}^\star
= \arg\max_{\mathbf{H} \in \mathcal{H}}
R\!\left(f_{\mathrm{gen}}(\mathbf{H}, \mathbf{x}, \mathbf{v}; \theta)\right) ,
\label{eq:latent-opt}
\end{equation}
where $\mathcal{H}$ denotes the space of latent trajectories with dimensions compatible with the pretrained model, and $f_{\mathrm{gen}}$ performs deterministic decoding given a fixed latent sequence. The optimization operates solely on $\mathbf{H}$ while keeping $\theta$ fixed, enabling reward-guided refinement at test time without retraining. The fixed generator $f_{\mathrm{gen}}$ defines how a candidate latent sequence is decoded; the quadratic anchoring term in Eq.~\eqref{eq:segment-loss} penalizes deviations from the initial backbone-induced trajectory. Algorithm~\ref{alg:ourmethod} applies this objective sequentially to each active latent segment.

\subsection{Latent Weaving: Sequential Segment Optimization}
\label{latent-weave}
Building on the latent optimization objective in Eq.~\eqref{eq:latent-opt}, we introduce a structured strategy to optimize long latent trajectories efficiently and stably. The same segment procedure is used in the offline collection stage and in evaluation-time refinement (Phases~I and III of Algorithm~\ref{alg:ourmethod}), with ground-truth reward restricted to offline collection and verifier-only reward used at evaluation.

\paragraph{Structured Latent Optimization.}
Optimizing the entire latent sequence $\mathbf{H}$ couples many decoding positions and can be costly, while early perturbations may propagate through later generation. We therefore propose \emph{Latent Weaving}, which decomposes latent optimization into localized, sequential segments. This decomposition allows each portion of the latent trajectory to be refined while leaving the reasoning states in all previously accepted segments unchanged.

Specifically, we partition the generation timeline into contiguous segments $\mathcal{S}_i$:
\begin{equation}
\mathcal{S}_i = \{s_i, s_i+1, \dots, e_i\}, \qquad
|\mathcal{S}_i| = \min\!\left\{\max(1,\lfloor\rho T\rfloor),\,T-s_i+1\right\} ,
\label{eq:segment}
\end{equation}
where $T$ is the current generation length and $\rho \in (0,1)$ controls the nominal segment length. Contiguous segments cover the trajectory in temporal order, with $s_1=1$, $s_{i+1}=e_i+1$, $N_{\mathrm{seg}}=\lceil T/\max(1,\lfloor\rho T\rfloor)\rceil$, and a possibly shorter final segment. Appendix~\ref{sec:appendix_optimization_protocol} specifies the four-segment configuration used in the component comparison.

\paragraph{Segment-wise Optimization Objective.}
For each segment, only the corresponding hidden states receive gradient updates; the accepted prefix remains fixed, and the suffix is regenerated after the update. The optimization objective uses a score-function estimator \citep{williams1992simple} with a reward baseline, following the sequence-level training principle of \citet{rennie2017self}:
\begin{equation}
\mathcal{L}_{\mathrm{opt}}(\mathbf{H}_{\mathcal{S}_i})
= -\frac{1}{M}\sum_{m=1}^{M} A^{(m)}
\sum_{t \in \mathcal{S}_i}
\log p_\theta\!\left(y_t^{(m)} \mid \mathbf{h}_t^{(m)}\right)
+\frac{\lambda}{2}
\left\|\mathbf{H}_{\mathcal{S}_i}-\mathbf{H}^{(0)}_{\mathcal{S}_i}\right\|_2^2 ,
\label{eq:segment-loss}
\end{equation}
where $M\geq2$ stochastic inner-loop rollouts are sampled at temperature $\tau$, and $A^{(m)}=R(\mathbf{y}^{(m)})-b^{(m)}$ is a centered advantage. The leave-one-out baseline is $b^{(m)}=(M-1)^{-1}\sum_{n\neq m}R(\mathbf{y}^{(n)})$. The superscript $m$ indexes a rollout and its hidden states. Rewards and baselines are treated as scalar feedback when differentiating the log probabilities; gradients update the active latent variables while the backbone and verifier remain fixed. The regularizer anchors the optimized states to the backbone-induced trajectory. Stochastic rollouts supply the optimization signal, whereas the final benchmark answer is decoded deterministically. After each inner update, the suffix beginning at the active segment is re-decoded autoregressively, while the prefix cache and previously accepted segments remain fixed. Appendix~\ref{sec:appendix_optimization_protocol} reports $M$, $\tau$, and $\lambda$ and the re-decoding policy.

The hidden states are updated by gradient descent on Eq.~\eqref{eq:segment-loss}:
\begin{equation}
\mathbf{H}_{\mathcal{S}_i}^{(k+1)}
= \mathbf{H}_{\mathcal{S}_i}^{(k)}
- \eta \nabla_{\mathbf{H}_{\mathcal{S}_i}}
\mathcal{L}_{\mathrm{opt}}\!\left(\mathbf{H}_{\mathcal{S}_i}^{(k)}\right) .
\label{eq:segment-update}
\end{equation}

\paragraph{Sequential Weaving.}
Segments are optimized in temporal order. When optimizing the $i$-th segment, all previously optimized segments are fixed and act as latent context. This freezing mechanism preserves accepted prefix states, although the reward of the completed answer can still change as the suffix is refined. Formally, the active segment is optimized under the following fixed-prefix constraint:
\begin{equation}
\mathbf{H}_{\mathcal{S}_i}^\star
= \arg\max_{\mathbf{H}_{\mathcal{S}_i}}
R\!\left(
f_{\mathrm{gen}}\big(
\mathbf{H}_{\mathcal{S}_{<i}}^\star,
\mathbf{H}_{\mathcal{S}_i},
\mathbf{H}_{\mathrm{ctx},i}; \theta
\big)
\right) ,
\label{eq:sequential}
\end{equation}
where $\mathbf{H}_{\mathcal{S}_{<i}}^\star$ denotes previously optimized segments and $\mathbf{H}_{\mathrm{ctx},i}$ denotes the remaining states, which receive no direct gradient update in the current step. The regenerated continuation can nevertheless change through autoregressive decoding. The input $(\mathbf{x},\mathbf{v})$ is implicit in this expression. Algorithm~\ref{alg:ourmethod} summarizes the segment-wise freezing and regeneration schedule.

\definecolor{down}{rgb}{0.8, 0, 0.0}
\definecolor{up}{rgb}{0.0, 0.7, 0.0}
\begin{table}[!t]

\centering

\footnotesize
\setlength{\tabcolsep}{2pt}

\caption{\textbf{Main results} under two MLLM backbones across mathematical reasoning, visual QA, and diagram understanding. The best and second-best results per benchmark are shown in \textbf{bold} and \underline{underline}, respectively.}
\begin{tabular}{l ccc ccc c}

\toprule

\multirow{2}{*}{\textbf{Method}} &

\multicolumn{3}{c}{\textbf{Mathematical Reasoning}} &

\multicolumn{3}{c}{\textbf{General Visual QA}} &

\textbf{Diagram} \\

\cmidrule(lr){2-4} \cmidrule(lr){5-7} \cmidrule(lr){8-8}

 & \shortstack{Math\\Vision} & \shortstack{Math\\Verse} & \shortstack{Math\\Vista} & MMMU & \shortstack{RealWorld\\QA} & MME & AI2D \\

\midrule

\multicolumn{8}{c}{\textbf{\llmname{InternVL2.5-4B}}} \\

\midrule

Vanilla             & $19.58$ & $34.49$ & $60.17$ & $46.82$ & $61.04$ & $2316$ & $79.63$ \\

CoT                 &$20.93_{\textcolor{up}{\uparrow 1.35}}$
                    &$35.95_{\textcolor{up}{\uparrow 1.46}}$
                    &$60.81_{\textcolor{up}{\uparrow 0.64}}$
                    &$48.22_{\textcolor{up}{\uparrow 1.40}}$
                    &$62.19_{\textcolor{up}{\uparrow 1.15}}$
                    &$2331_{\textcolor{up}{\uparrow 15.0}}$
                    &$80.10_{\textcolor{up}{\uparrow 0.47}}$\\

\cmidrule(lr){1-8}

SFT                 &$19.28_{\textcolor{down}{\downarrow 0.30}}$
                    &$35.13_{\textcolor{up}{\uparrow 0.64}}$
                    &$60.82_{\textcolor{up}{\uparrow 0.65}}$
                    &$49.51_{\textcolor{up}{\uparrow 2.69}}$
                    &$63.29_{\textcolor{up}{\uparrow 2.25}}$
                    &$2309_{\textcolor{down}{\downarrow 7.0}}$
                    &$80.87_{\textcolor{up}{\uparrow 1.24}}$\\

GRPO                &$20.07_{\textcolor{up}{\uparrow 0.49}}$
                    &$35.66_{\textcolor{up}{\uparrow 1.17}}$
                    &$61.34_{\textcolor{up}{\uparrow 1.17}}$
                    &$48.86_{\textcolor{up}{\uparrow 2.04}}$
                    &$61.35_{\textcolor{up}{\uparrow 0.31}}$
                    &$2302_{\textcolor{down}{\downarrow 14.0}}$
                    &$78.41_{\textcolor{down}{\downarrow 1.22}}$\\

Self-Rewarding      &$21.32_{\textcolor{up}{\uparrow 1.74}}$
                    &$36.17_{\textcolor{up}{\uparrow 1.68}}$
                    &$60.67_{\textcolor{up}{\uparrow 0.50}}$
                    &$49.42_{\textcolor{up}{\uparrow 2.60}}$
                    &$62.90_{\textcolor{up}{\uparrow 1.86}}$
                    &$2315_{\textcolor{down}{\downarrow 1.0}}$
                    &$81.04_{\textcolor{up}{\uparrow 1.41}}$\\

\cmidrule(lr){1-8}

IVT-LR\textsuperscript{$\dagger$} &$22.31_{\textcolor{up}{\uparrow 2.73}}$
                    &$36.19_{\textcolor{up}{\uparrow 1.70}}$
                    &$61.20_{\textcolor{up}{\uparrow 1.03}}$
                    &$49.68_{\textcolor{up}{\uparrow 2.86}}$
                    &$\underline{66.41}_{\textcolor{up}{\uparrow 5.37}}$
                    &$2333_{\textcolor{up}{\uparrow 17.0}}$
                    &$82.12_{\textcolor{up}{\uparrow 2.49}}$\\

LaCoT {\scriptsize(\texttt{NeurIPS} \citeyear{sun2025latent})} &$20.87_{\textcolor{up}{\uparrow 1.29}}$
                    &$\underline{41.23}_{\textcolor{up}{\uparrow 6.74}}$
                    &$59.34_{\textcolor{down}{\downarrow 0.83}}$
                    &$50.45_{\textcolor{up}{\uparrow 3.63}}$
                    &$64.28_{\textcolor{up}{\uparrow 3.24}}$
                    &$2326_{\textcolor{up}{\uparrow 10.0}}$
                    &$81.52_{\textcolor{up}{\uparrow 1.89}}$\\

\cmidrule(lr){1-8}

Self-Refine         &$23.38_{\textcolor{up}{\uparrow 3.80}}$
                    &$38.88_{\textcolor{up}{\uparrow 4.39}}$
                    &$62.72_{\textcolor{up}{\uparrow 2.55}}$
                    &$50.91_{\textcolor{up}{\uparrow 4.09}}$
                    &$62.78_{\textcolor{up}{\uparrow 1.74}}$
                    &$2331_{\textcolor{up}{\uparrow 15.0}}$
                    &$\underline{82.48}_{\textcolor{up}{\uparrow 2.85}}$\\

Self-Consistency    &$23.06_{\textcolor{up}{\uparrow 3.48}}$
                    &$37.89_{\textcolor{up}{\uparrow 3.40}}$
                    &$61.62_{\textcolor{up}{\uparrow 1.45}}$
                    &$51.02_{\textcolor{up}{\uparrow 4.20}}$
                    &$63.08_{\textcolor{up}{\uparrow 2.04}}$
                    &$2329_{\textcolor{up}{\uparrow 13.0}}$
                    &$81.05_{\textcolor{up}{\uparrow 1.42}}$\\

TTRV {\scriptsize(\texttt{CVPR} \citeyear{singh2026ttrv})} &$\underline{25.30}_{\textcolor{up}{\uparrow 5.72}}$
                    &$40.12_{\textcolor{up}{\uparrow 5.63}}$
                    &$\underline{64.14}_{\textcolor{up}{\uparrow 3.97}}$
                    &$\mathbf{52.29}_{\textcolor{up}{\uparrow 5.47}}$
                    &$66.02_{\textcolor{up}{\uparrow 4.98}}$
                    &$\underline{2338}_{\textcolor{up}{\uparrow 22.0}}$
                    &$82.17_{\textcolor{up}{\uparrow 2.54}}$\\

\textbf{\ourmethod} &$\mathbf{26.59}_{\textcolor{up}{\uparrow 7.01}}$
                    &$\mathbf{43.22}_{\textcolor{up}{\uparrow 8.73}}$
                    &$\mathbf{66.19}_{\textcolor{up}{\uparrow 6.02}}$
                    &$\underline{52.14}_{\textcolor{up}{\uparrow 5.32}}$
                    &$\mathbf{67.80}_{\textcolor{up}{\uparrow 6.76}}$
                    &$\mathbf{2345}_{\textcolor{up}{\uparrow 29.0}}$
                    &$\mathbf{84.64}_{\textcolor{up}{\uparrow 5.01}}$\\

\midrule

\multicolumn{8}{c}{\textbf{\llmname{Qwen2.5-VL-7B}}} \\

\midrule

Vanilla             & $23.65$ & $47.98$ & $66.15$ & $56.96$ & $66.02$ & $2328$ & $80.58$ \\

CoT                 &$24.38_{\textcolor{up}{\uparrow 0.73}}$
                    &$48.66_{\textcolor{up}{\uparrow 0.68}}$
                    &$66.91_{\textcolor{up}{\uparrow 0.76}}$
                    &$57.92_{\textcolor{up}{\uparrow 0.96}}$
                    &$67.35_{\textcolor{up}{\uparrow 1.33}}$
                    &$2344_{\textcolor{up}{\uparrow 16.0}}$
                    &$82.98_{\textcolor{up}{\uparrow 2.40}}$\\

\cmidrule(lr){1-8}

SFT                 &$22.81_{\textcolor{down}{\downarrow 0.84}}$
                    &$47.03_{\textcolor{down}{\downarrow 0.95}}$
                    &$66.13_{\textcolor{down}{\downarrow 0.02}}$
                    &$57.18_{\textcolor{up}{\uparrow 0.22}}$
                    &$65.95_{\textcolor{down}{\downarrow 0.07}}$
                    &$2338_{\textcolor{up}{\uparrow 10.0}}$
                    &$81.49_{\textcolor{up}{\uparrow 0.91}}$\\

GRPO                &$23.69_{\textcolor{up}{\uparrow 0.04}}$
                    &$48.55_{\textcolor{up}{\uparrow 0.57}}$
                    &$67.70_{\textcolor{up}{\uparrow 1.55}}$
                    &$58.61_{\textcolor{up}{\uparrow 1.65}}$
                    &$67.36_{\textcolor{up}{\uparrow 1.34}}$
                    &$2340_{\textcolor{up}{\uparrow 12.0}}$
                    &$81.65_{\textcolor{up}{\uparrow 1.07}}$\\

Self-Rewarding      &$23.92_{\textcolor{up}{\uparrow 0.27}}$
                    &$48.47_{\textcolor{up}{\uparrow 0.49}}$
                    &$68.14_{\textcolor{up}{\uparrow 1.99}}$
                    &$59.44_{\textcolor{up}{\uparrow 2.48}}$
                    &$66.99_{\textcolor{up}{\uparrow 0.97}}$
                    &$2356_{\textcolor{up}{\uparrow 28.0}}$
                    &$82.12_{\textcolor{up}{\uparrow 1.54}}$\\

\cmidrule(lr){1-8}

IVT-LR\textsuperscript{$\dagger$} &$26.36_{\textcolor{up}{\uparrow 2.71}}$
                    &$49.48_{\textcolor{up}{\uparrow 1.50}}$
                    &$67.26_{\textcolor{up}{\uparrow 1.11}}$
                    &$59.43_{\textcolor{up}{\uparrow 2.47}}$
                    &$69.30_{\textcolor{up}{\uparrow 3.28}}$
                    &$2351_{\textcolor{up}{\uparrow 23.0}}$
                    &$84.02_{\textcolor{up}{\uparrow 3.44}}$\\

LaCoT {\scriptsize(\texttt{NeurIPS} \citeyear{sun2025latent})} &$\underline{27.85}_{\textcolor{up}{\uparrow 4.20}}$
                    &$48.70_{\textcolor{up}{\uparrow 0.72}}$
                    &$67.59_{\textcolor{up}{\uparrow 1.44}}$
                    &$58.32_{\textcolor{up}{\uparrow 1.36}}$
                    &$70.14_{\textcolor{up}{\uparrow 4.12}}$
                    &$2368_{\textcolor{up}{\uparrow 40.0}}$
                    &$83.31_{\textcolor{up}{\uparrow 2.73}}$\\

\cmidrule(lr){1-8}

Self-Refine         &$24.40_{\textcolor{up}{\uparrow 0.75}}$
                    &$49.53_{\textcolor{up}{\uparrow 1.55}}$
                    &$68.16_{\textcolor{up}{\uparrow 2.01}}$
                    &$57.97_{\textcolor{up}{\uparrow 1.01}}$
                    &$68.20_{\textcolor{up}{\uparrow 2.18}}$
                    &$2355_{\textcolor{up}{\uparrow 27.0}}$
                    &$\underline{86.30}_{\textcolor{up}{\uparrow 5.72}}$\\

Self-Consistency    &$25.01_{\textcolor{up}{\uparrow 1.36}}$
                    &$50.82_{\textcolor{up}{\uparrow 2.84}}$
                    &$\underline{69.78}_{\textcolor{up}{\uparrow 3.63}}$
                    &$58.57_{\textcolor{up}{\uparrow 1.61}}$
                    &$67.96_{\textcolor{up}{\uparrow 1.94}}$
                    &$2358_{\textcolor{up}{\uparrow 30.0}}$
                    &$84.94_{\textcolor{up}{\uparrow 4.36}}$\\

TTRV {\scriptsize(\texttt{CVPR} \citeyear{singh2026ttrv})} &$27.08_{\textcolor{up}{\uparrow 3.43}}$
                    &$\underline{51.76}_{\textcolor{up}{\uparrow 3.78}}$
                    &$69.26_{\textcolor{up}{\uparrow 3.11}}$
                    &$\underline{61.40}_{\textcolor{up}{\uparrow 4.44}}$
                    &$\mathbf{71.28}_{\textcolor{up}{\uparrow 5.26}}$
                    &$\underline{2375}_{\textcolor{up}{\uparrow 47.0}}$
                    &$86.21_{\textcolor{up}{\uparrow 5.63}}$\\

\textbf{\ourmethod} &$\mathbf{30.11}_{\textcolor{up}{\uparrow 6.46}}$
                    &$\mathbf{53.80}_{\textcolor{up}{\uparrow 5.82}}$
                    &$\mathbf{72.03}_{\textcolor{up}{\uparrow 5.88}}$
                    &$\mathbf{63.15}_{\textcolor{up}{\uparrow 6.19}}$
                    &$\underline{70.89}_{\textcolor{up}{\uparrow 4.87}}$
                    &$\mathbf{2388}_{\textcolor{up}{\uparrow 60.0}}$
                    &$\mathbf{88.93}_{\textcolor{up}{\uparrow 8.35}}$\\

\bottomrule
\end{tabular}
\label{tab:main_results}
\end{table}

\subsection{Optimization Signals for Latent Weaving}\label{reward}
We describe the reward signals for latent weaving. During offline trajectory collection (Phase~I), the declared cold-start term may be used on the training split; during evaluation-time refinement (Phase~III), only verifier rewards are permitted. In both cases, the scalar feedback drives gradient-based updates of the active segment while the backbone remains fixed.

\paragraph{Reward-Guided Control.}
The reward combines structural verification with a weak correctness signal during offline collection, and uses verification alone during evaluation:
\begin{equation}
R(\mathbf{y}) =
\begin{cases}
\alpha\, R_{\mathrm{ver}}(\mathbf{y}, \mathbf{x}, \mathbf{v})
+ \beta\, R_{\mathrm{gt}}(\mathbf{y}, \mathbf{y}^\star),
& \text{cold-start}, \\[4pt]
\alpha\, R_{\mathrm{ver}}(\mathbf{y}, \mathbf{x}, \mathbf{v}),
& \text{test-time},
\end{cases}
\label{eq:reward}
\end{equation}
where $\alpha$ scales verifier feedback and $\beta$ controls the ground-truth term. Cold-start collection uses disjoint offline training data; development and test evaluation set $\beta=0$ and use verifier feedback alone. Appendix~\ref{sec:appendix_reward_config} specifies the verifier composition and reward weights.

\paragraph{Verification Reward.}
The verifier reward aggregates four complementary checks: reasoning consistency, final-answer correctness, answer completeness, and question understanding. The correctness check judges the proposed answer from the input and visual evidence; the separate ground-truth term compares against a reference answer during offline collection. With nonnegative weights and a positive total weight, the normalized verifier reward lies in $[-1,0]$, and a larger value indicates fewer failed checks:
\begin{equation}
R_{\mathrm{ver}}
= - \frac{
\sum_{k \in \mathcal{F}} w_k \, \mathbb{1}[k(\mathbf{y}) = 0]
}{
\sum_{k \in \mathcal{F}} w_k
} ,
\label{eq:verifier}
\end{equation}
where $\mathcal{F}$ is a predefined set of fixed verification functions for the current task, and $w_k$ denotes the relative importance of each individual verifier. These verifiers provide structured, task-specific feedback without updating the evaluated backbone parameters. Appendix~\ref{sec:appendix_verifier_prompts} lists the complete prompt templates for each of the four task-specific verifier checks used during optimization.

\paragraph{Ground-Truth Penalty.}
The auxiliary correctness signal is defined as:
\begin{equation}
R_{\mathrm{gt}}(\mathbf{y}, \mathbf{y}^\star)
= \mathbb{1}[\operatorname{Ans}(\mathbf{y}) \neq \mathbf{y}^\star] \cdot (-1) ,
\label{eq:gt}
\end{equation}
where $\operatorname{Ans}$ extracts the benchmark-formatted final answer from the generated reasoning trace. An answer mismatch incurs $-1$, and a match incurs $0$; this signal is used only on the offline training split.

\begingroup
\renewcommand{\thefootnote}{\fnsymbol{footnote}}
\footnotetext[2]{IVT-LR: ModalityDance Research, 2026-1-29 (\href{https://github.com/ModalityDance/IVT-LR}{project repository}).}
\endgroup

\subsection{Reasoning Anchors for Latent Weaving}\label{anchors}
Finally, we introduce \emph{Reasoning Anchors}, an offline-trained mechanism that complements test-time latent weaving with explicit reasoning structure. This procedure corresponds to Phase~II in Algorithm~\ref{alg:ourmethod} and is applied after collecting optimized trajectories on the training split. The trajectories are converted to textual reasoning traces and bound to explicit \texttt{<WEAVE>} delimiters. At inference, WEAVEGen produces one anchor sequence from $(\mathbf{x},\mathbf{v})$; that text is inserted into the evaluated backbone's context before latent weaving. The interface is textual, so WEAVEGen and the evaluated backbone need not share hidden size or parameters.

\paragraph{WEAVEGen Model.}
WEAVEGen is a compact auxiliary generator parameterized by $\phi$. It is initialized from \llmname{Qwen2-VL-2B-Instruct} \citep{wang2024qwen2vl} and trained separately to generate structured textual reasoning. Building on explicit intermediate-step supervision \citep{wei2022chain, nye2021show}, we serialize each step between \texttt{<WEAVE>} delimiters. This 2B auxiliary model has its own parameters and tokenizer; its generated text is consumed through the ordinary language-token interface of the evaluated 4B--27B backbone $f_\theta$. Appendix~\ref{sec:appendix_efp} accounts for the tokens, latency, and memory of its single auxiliary generation call.

\begin{equation}
\mathcal{L}_{\mathrm{bind}}
= - \sum_{j=1}^{J}
\log p_\phi\!\left(
\mathrm{step}_j \mid \mathbf{x}, \mathbf{v}, \mathbf{a}_{<j}
\right) ,
\label{eq:bind}
\end{equation}
where $J$ is the number of anchor steps and $\mathbf{a}_{<j}$ is their serialized text prefix, including \texttt{<WEAVE>} delimiters. The step likelihood under WEAVEGen, $p_\phi$, is factorized over the tokens of $\mathrm{step}_j$. Here $J$ is distinct from the rollout count $M$ and the update count $K$.

To encourage WEAVEGen to encode answer-relevant information in its anchors, we introduce an auxiliary answer-prediction objective:
\begin{equation}
\mathcal{L}_{\mathrm{util}}
= - \mathbb{E}
\log p_\phi\!\left(
\mathbf{y}^\star \mid \mathbf{x}, \mathbf{v},
\mathbf{a}
\right) .
\label{eq:util}
\end{equation}
Here $\mathbf{a}$ is the serialized anchor sequence and the expectation is over offline training examples and their anchor targets. Together, the objectives train WEAVEGen to emit a structured textual scaffold that conditions the backbone before latent optimization, without transferring hidden states. Appendix~\ref{sec:appendix_anchor_protocol} specifies serialization, insertion position, anchor length, and caching.

\section{Experiments}

In this section, we conduct extensive experiments to evaluate the effectiveness of \ourmethod for multimodal reasoning, organized around explicit research questions to analyze component contributions and reasoning behaviors.
Specifically, we aim to answer the following research questions:
\begin{itemize}[leftmargin=*,before=\interlinepenalty10000,beginpenalty=10000,midpenalty=10000]
    \item \textbf{RQ1.} How well does \ourmethod improve multimodal reasoning compared to baselines?
    \item \textbf{RQ2.} How do Latent Weaving and Reasoning Anchors affect \ourmethod's performance?
    \item \textbf{RQ3.} How sensitive is \ourmethod to key design choices and hyperparameters?
\end{itemize}

\subsection{Experiment Setup}
\label{setup}
\paragraph{Tasks \& Benchmarks.} We evaluate three task categories: \ding{117} \textit{Mathematical reasoning}, covering MathVision \citep{wang2024measuring}, MathVerse \citep{zhang2024mathverse}, and MathVista \citep{lu2023mathvista}; \ding{117} \textit{General Visual QA}, including MMMU \citep{yue2024mmmu}, RealWorldQA \citep{xai2024grok15v}, and MME \citep{fu2023mme}; \ding{117} \textit{Diagram Understanding}, evaluated on AI2D \citep{kembhavi2016diagram}.
We report accuracy (\%) for all benchmarks except MME, which uses its aggregate benchmark score. Accordingly, accuracy gains are expressed in percentage points and MME gains in score points. Appendix~\ref{sec:appendix_dataset_stats} lists the evaluation splits and dataset sizes.

\begin{table}[!t]
\centering
\caption{\textbf{Performance gains} of \ourmethod over vanilla decoding across five MLLM backbones on diverse multimodal reasoning benchmarks. $\Delta$ denotes the absolute improvement on each benchmark.}
\label{tab:multi-mllm}
\small
\setlength{\tabcolsep}{3pt}
\resizebox{\columnwidth}{!}{
\begin{tabular}{@{}llccccccc@{}}
\toprule
\textbf{Backbone} & \textbf{Method}
& \textbf{MathVision}
& \textbf{MathVerse}
& \textbf{MathVista}
& \textbf{MMMU}
& \textbf{RealWorldQA}
& \textbf{MME}
& \textbf{AI2D} \\
\midrule

\multirow{3}{*}{\llmname{Qwen2.5-VL-7B}}
 & Vanilla      & 23.65 & 47.98 & 66.15 & 56.96 & 66.02 & 2328 & 80.58 \\
 & +\ourmethod  & 30.11 & 53.80 & 72.03 & 63.15 & 70.89 & 2388 & 88.93  \\
 & $\Delta$     & \textcolor{brown}{+6.46} & \textcolor{brown}{+5.82} & \textcolor{brown}{+5.88} & \textcolor{brown}{+6.19} & \textcolor{brown}{+4.87} & \textcolor{brown}{+60.0} & \textcolor{brown}{+8.35} \\
\midrule

\multirow{3}{*}{\llmname{Qwen3.5-4B}}
 & Vanilla & 66.12 & 66.37 & 75.80 & 68.32 & 73.33 & 2372 & 85.91 \\
 & +\ourmethod & 72.69 & 71.44 & 79.30 & 72.81 & 76.71 & 2404 & 89.25 \\
 & $\Delta$ & \textcolor{brown}{+6.57} & \textcolor{brown}{+5.07} & \textcolor{brown}{+3.50} & \textcolor{brown}{+4.49} & \textcolor{brown}{+3.38} & \textcolor{brown}{+32.0} & \textcolor{brown}{+3.34} \\
\midrule

\multirow{3}{*}{\llmname{Qwen3.6-27B}}
 & Vanilla & 75.33 & 74.36 & 82.30 & 72.56 & 77.12 & 2456 & 90.31 \\
 & +\ourmethod & 81.90 & 78.04 & 85.90 & 76.38 & 82.34 & 2482 & 92.48 \\
 & $\Delta$ & \textcolor{brown}{+6.57} & \textcolor{brown}{+3.68} & \textcolor{brown}{+3.60} & \textcolor{brown}{+3.82} & \textcolor{brown}{+5.22} & \textcolor{brown}{+26.0} & \textcolor{brown}{+2.17} \\
\midrule

\multirow{3}{*}{\llmname{InternVL2.5-4B}}
 & Vanilla      & 19.58 & 34.49 & 60.17 & 46.82 & 61.04 & 2316 & 79.63 \\
 & +\ourmethod  & 26.59 & 43.22 & 66.19 & 52.14 & 67.80 & 2345 & 84.64  \\
 & $\Delta$     & \textcolor{brown}{+7.01} & \textcolor{brown}{+8.73} & \textcolor{brown}{+6.02} & \textcolor{brown}{+5.32} & \textcolor{brown}{+6.76} & \textcolor{brown}{+29.0} & \textcolor{brown}{+5.01} \\
\midrule

\multirow{3}{*}{\llmname{STEP3-VL-10B}}
 & Vanilla      & 68.77 & 67.20 & 80.59 & 75.12 & 73.45 & 2432 & 87.29 \\
 & +\ourmethod  & 70.29 & 72.33 & 81.98 & 78.64 & 77.70 & 2440 & 89.93 \\
 & $\Delta$     & \textcolor{brown}{+1.52} & \textcolor{brown}{+5.13} & \textcolor{brown}{+1.39} & \textcolor{brown}{+3.52} & \textcolor{brown}{+4.25} & \textcolor{brown}{+8.0} & \textcolor{brown}{+2.64} \\

\bottomrule
\end{tabular}
}
\end{table}

\paragraph{Baselines.} \ourmethod is compared against key methods across different reasoning paradigms:
\begin{itemize}[leftmargin=1em]
\item \textbf{Prompt-based Inference}: Vanilla decoding and Chain-of-Thought (CoT) \citep{wei2022chain}.
\item \textbf{Learning-based Optimization}:
supervised fine-tuning (SFT), Group Relative Policy Optimization (GRPO) \citep{shao2024deepseekmath}, and Self-Rewarding \citep{yuan2025selfrewardinglanguagemodels}.
\item \textbf{Latent-space Reasoning}:
methods operating on implicit reasoning representations, including IVT-LR \citep{chen2025reasoning} and LaCoT \citep{sun2025latent}.
\item \textbf{Test-time Scaling and Adaptation}:
sampling and refinement through Self-Consistency \citep{wang2023selfconsistencyimproveschainthought}, Self-Refine \citep{madaan2023selfrefineiterativerefinementselffeedback}, and same-verifier Best-of-$N$; and test-time parameter updates through TTRV \citep{singh2026ttrv}, which uses frequency- and entropy-based rewards.
\end{itemize}

\paragraph{Backbones.} To assess \ourmethod's generalization capability, we conduct experiments on MLLMs drawn from several diverse model families, including Qwen2.5-VL-7B-Instruct \citep{bai2025qwen2}, Qwen3.5-4B \citep{qwen2026qwen35}, Qwen3.6-27B \citep{qwen2026qwen36}, InternVL2.5-4B \citep{chen2024expanding}, and STEP3-VL-10B \citep{huang2026step3}.

\paragraph{Implementation details.} Final benchmark answers use deterministic decoding with maximum length 1024. Inner-loop optimization instead uses the stochastic estimator in Eq.~\eqref{eq:segment-loss}; its rollout count, temperature, baseline, clipping, regularization, segment coverage, update count, and suffix re-decoding policy are reported per configuration in Appendix~\ref{sec:appendix_optimization_protocol}. Ground-truth reward is restricted to offline trajectory collection on a training split and is disabled for every evaluation example. WEAVEGen is trained offline with $\mathcal{L}_{\mathrm{bind}}+\mathcal{L}_{\mathrm{util}}$ and supplies textual anchors through the interface in Appendix~\ref{sec:appendix_anchor_protocol}. In the four-segment component comparison, Best-of-$N$ and the latent variants each use 96 candidate-level verifier evaluations; their tokens, latency, and peak memory are reported in Appendix~\ref{sec:appendix_efp}. Experiments use one machine with 8 NVIDIA A100 (80GB) GPUs.

\begin{figure}[!t]
    \centering
    \includegraphics[width=\linewidth]{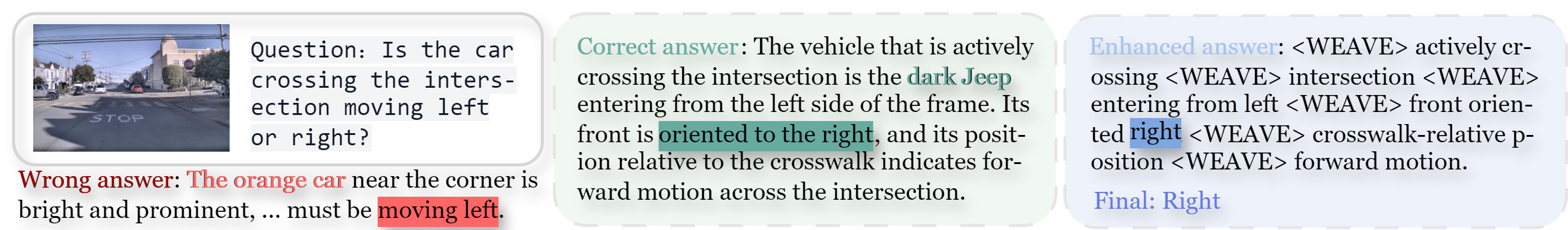}
    \caption{Qualitative case study of reasoning error accumulation and correction. \textbf{Left}: Incorrect early reasoning leads to error propagation. \textbf{Middle}: Latent Weaving corrects the reasoning trajectory. \textbf{Right}: \texttt{<WEAVE>} tokens enhance the clarity and structure of the corrected reasoning trajectory.}
    \label{fig:case_study}
\end{figure}

\subsection{Performance and Main Results (RQ1)}

\paragraph{Consistent improvements across task domains.}
Table~\ref{tab:main_results} shows that \ourmethod improves over vanilla decoding across all evaluated task categories. With \llmname{InternVL2.5-4B}, accuracy increases by $7.01$, $8.73$, and $6.02$ percentage points on MathVision, MathVerse, and MathVista, respectively, and by $5.32$, $6.76$, and $5.01$ points on MMMU, RealWorldQA, and AI2D. With \llmname{Qwen2.5-VL-7B}, all three mathematical benchmarks improve by more than five percentage points, alongside gains of $6.19$ points on MMMU and $8.35$ on AI2D. The method improves every evaluated benchmark over its corresponding vanilla baseline, although individual competing methods remain stronger on some tasks, such as TTRV on Qwen2.5-VL-7B RealWorldQA.

\paragraph{Robust generalization across model backbones.}
As summarized in Table~\ref{tab:multi-mllm}, \ourmethod improves all seven benchmarks across five MLLM backbones spanning 4B to 27B parameters. On MathVision, gains range from $+1.52$ percentage points (\llmname{STEP3-VL-10B}) to $+7.01$ (\llmname{InternVL2.5-4B}), with $+6.57$ for \llmname{Qwen3.5-4B} and $+6.57$ for \llmname{Qwen3.6-27B}. Gains also remain positive on MMMU ($+3.52$ to $+6.19$ points), RealWorldQA ($+3.38$ to $+6.76$ points), and AI2D ($+2.17$ to $+8.35$ points), while MME increases by 8 to 60 score points. These results extend the observed benefits to newer backbones and a larger 27B model, supporting the applicability of latent trajectory refinement across the evaluated model families and scales.

\begin{figure}[!t]
\centering
\begin{minipage}[t]{0.48\linewidth}
    \vspace{0pt}
    \centering
    \includegraphics[width=\linewidth,trim=0bp 26bp 36bp 40bp,clip]{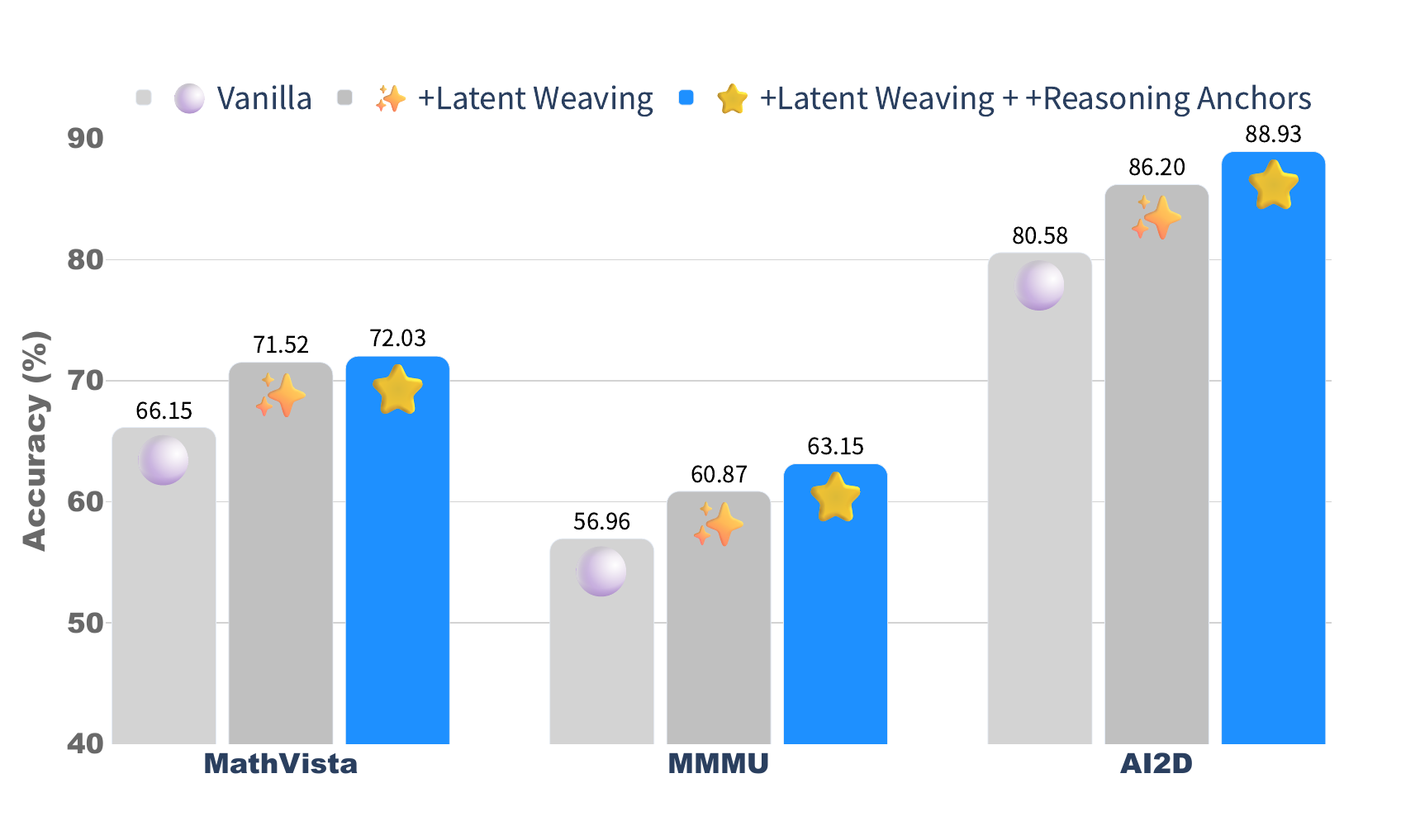}
    \caption{Component-wise performance comparison with Latent Weaving and Reasoning Anchors.}
    \label{fig:RQ2}
\end{minipage}\hfill
\begin{minipage}[t]{0.48\linewidth}
\vspace{0pt}
\centering
\small
\renewcommand{\arraystretch}{1.35}
\setlength{\tabcolsep}{2.5pt}
\captionof{table}{Latent Weaving by stage, measured in \emph{reasoning steps} of the generated chain of thought.}
\begin{tabular}[b]{lcc}
\toprule
\textbf{Weaving Steps} & \textbf{MathVista} & \textbf{MMMU} \\
\midrule
Vanilla              & 66.15 & 56.96 \\
Early (first 30\%)   & 67.02 & 57.41 \\
Middle only                  & 64.47 & 54.92 \\
Late (last 30\%) & 58.85 & 51.31 \\
\midrule
Full trajectory         & \textbf{72.03} & \textbf{63.15} \\
\bottomrule
\end{tabular}
\label{tab:partial_weaving}
\end{minipage}
\end{figure}

\paragraph{Effectiveness under strong baseline performance.}
The improvements achieved by \ourmethod persist even when Vanilla performance is already strong. On \llmname{Qwen3.6-27B}, MathVision rises from $75.33\%$ to $81.90\%$ ($+6.57$ percentage points), while RealWorldQA increases from $77.12\%$ to $82.34\%$ ($+5.22$ points). Even on AI2D, where Vanilla reaches $90.31\%$, \ourmethod achieves $92.48\%$ ($+2.17$ points). The smaller \llmname{Qwen3.5-4B} also improves from $66.12\%$ to $72.69\%$ on MathVision and from $75.80\%$ to $79.30\%$ on MathVista. Together with the gains on \llmname{STEP3-VL-10B}, these results indicate that the benefits are not confined to backbones with low initial accuracy, although the size of the gain varies across models and benchmarks.

\subsection{Component Contribution (RQ2)}

\paragraph{Effect of Latent Weaving.}
Figure~\ref{fig:RQ2} compares the component configurations. Latent Weaving alone reaches $71.52\%$ on MathVista, $60.87\%$ on MMMU, and $86.20\%$ on AI2D, improving on Vanilla across all three benchmarks. On MathVista, it exceeds the same-verifier Best-of-$N$ result of $69.44\%$ by $2.08$ percentage points. Table~\ref{tab:matched_verifier_control} reports both methods at a shared budget of 96 candidate-level verifier evaluations.

\paragraph{Complementary role of Reasoning Anchors.}
Reasoning Anchors are evaluated separately from latent updates. The $2\times2$ design in Table~\ref{tab:ablation_components} shows that adding anchors to latent updates raises MathVista from $71.52\%$ to $72.03\%$ ($+0.51$ points), MMMU from $60.87\%$ to $63.15\%$ ($+2.28$ points), and AI2D from $86.20\%$ to $88.93\%$ ($+2.73$ points). The full variant has the highest reported point estimate on all three benchmarks.

\paragraph{Accuracy and inference cost.}
The same-verifier comparison separates candidate-evaluation budget from execution cost. Best-of-$N$ selects among independently generated candidates, whereas Latent Weaving uses feedback to update the active segment before generating new suffixes. At the shared 96-candidate budget, the full method exceeds Best-of-$N$ by $2.59$ points on MathVista. Table~\ref{tab:appendix_efp} reports 52 versus 31 seconds per example and 41 versus 21 GB peak memory on the cost subset. These measurements include the full method's auxiliary anchor-generation call and quantify the resource cost associated with its accuracy gain.

\subsection{Sensitivity Analysis (RQ3)}
\label{RQ3}

\begingroup
\setlength{\intextsep}{0pt}
\begin{wrapfigure}{r}{0.5\textwidth}
    \centering
    \includegraphics[width=\linewidth]{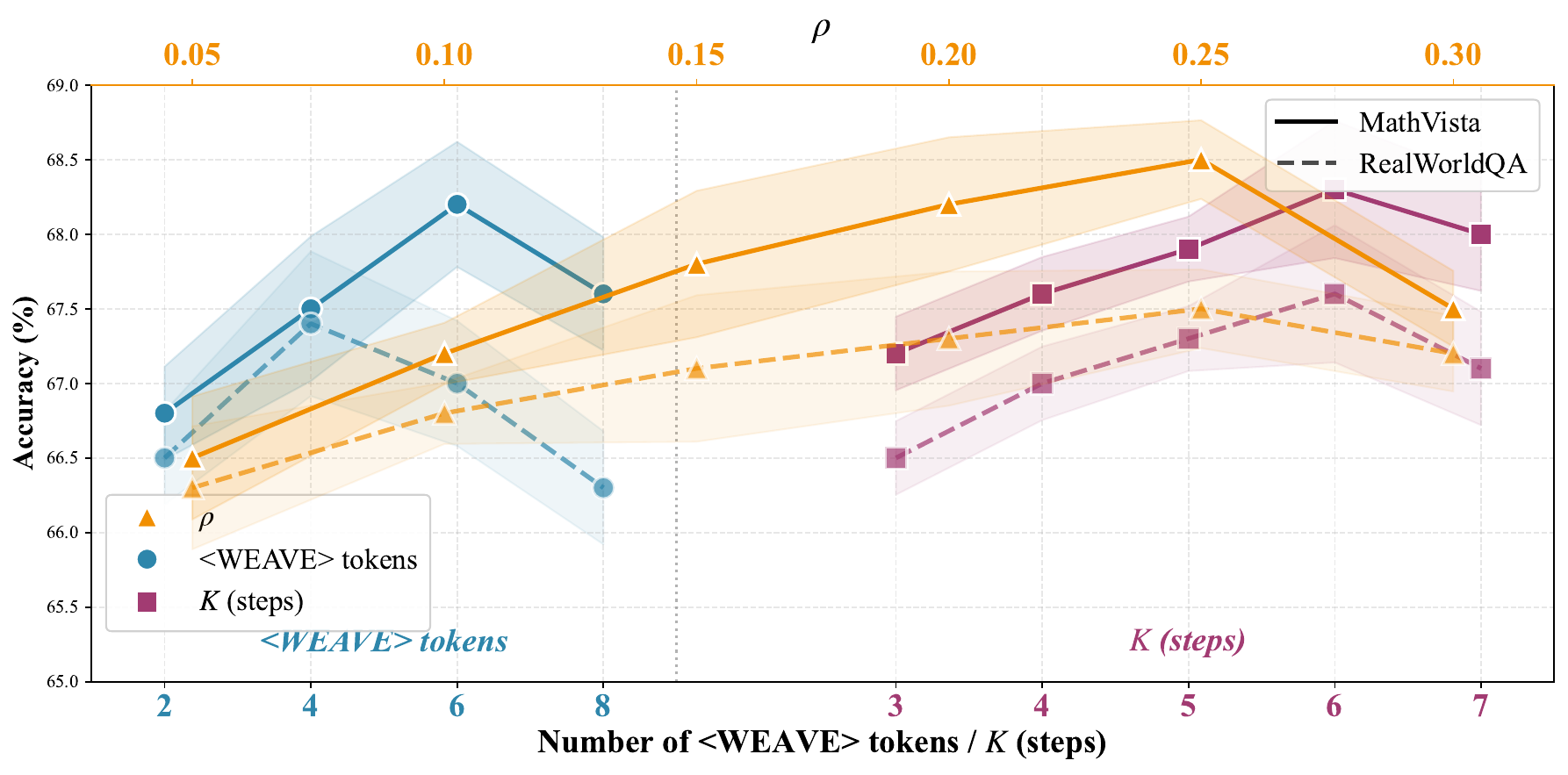}
    \caption{Sensitivity to segment ratio $\rho$, anchor-step count $J$ (one \texttt{<WEAVE>} opening delimiter per step), and update count $K$ on MathVista and RealWorldQA.}
    \label{fig:RQ3}
\end{wrapfigure}

We examine the segment ratio $\rho$, anchor count $J$, update count $K$, and verifier weights. Table~\ref{tab:ablation_weaving} transcribes the MathVista and RealWorldQA curves in Figure~\ref{fig:RQ3}, whose accuracy spreads are at most two percentage points. Each one-factor series compares settings within its reported sweep. The separate delimiter control in Table~\ref{tab:ablation_weaving} and the weight study in Table~\ref{tab:ablation_weights} report MathVista and MMMU. Hyperparameters are selected on a development split and frozen before test evaluation.

In addition to hyperparameter sensitivity, Table~\ref{tab:partial_weaving} compares intervention at different stages of a generated reasoning trajectory. These stages refer to reasoning steps, not the inner update count $K$. Early-only weaving improves MathVista and MMMU by $0.87$ and $0.45$ percentage points over vanilla decoding, whereas middle-only and late-only weaving reduce accuracy. Full-trajectory weaving performs best on both benchmarks, highlighting the role of early intervention in the sequential procedure. Additional ablations and sensitivity analyses are reported in Appendix~\ref{sec:appendix_ablation}.

\endgroup

\subsection{Case Study}
Figure~\ref{fig:case_study} illustrates how an early reasoning error affects the subsequent trace. The initial inference leads to a wrong final prediction (left), while Latent Weaving revises the active reasoning segment and recovers the correct answer (middle). The anchor-conditioned trace presents the corrected reasoning in explicit \texttt{<WEAVE>}-delimited steps (right). This example connects the latent intervention to the resulting textual continuation and illustrates the distinct roles of trajectory refinement and structured guidance.

\section{Conclusion}
In this paper, we study \ourmethod, a frozen-backbone test-time optimization framework that models multimodal reasoning as a controllable latent trajectory. Structured \textit{Latent Weaving} refines active hidden-state segments while \textit{Reasoning Anchors} provide textual guidance during generation. Across seven benchmarks and five backbones, the results support combining localized latent refinement with textual guidance. In the four-segment component comparison, Best-of-$N$ and the latent variants share the verifier and a 96-candidate evaluation budget, with resource costs reported separately. These results motivate direct control of the hidden-state trajectory that produces the final answer.

\medskip
\restorebackmatterspacing
\raggedbottom
\subsection*{AI use statement}

Writing assistance from language models was limited to polishing grammar and
wording in author-written prose and locating a subset of the papers discussed
in Related Work. AI coding assistants helped convert author-provided Excel
data into LaTeX format for Tables~\ref{tab:multi-mllm}
and~\ref{tab:partial_weaving}, and produce the bar chart in
Figure~\ref{fig:RQ2} and the line plot in Figure~\ref{fig:RQ3} from
author-provided data. AI tools were not used at any stage to generate research
ideas, define research objectives, run or interpret experiments, or generate
any reported numerical results. The authors retain responsibility for the
final manuscript.

\subsection*{Ethics statement}

This work studies test-time latent optimization for multimodal reasoning and is
conducted entirely on publicly available benchmarks (MathVision, MathVerse,
MathVista, MMMU, RealWorldQA, MME, and AI2D) under their released evaluation
protocols. No new data are collected from human subjects, and no sensitive,
private, or proprietary information is used. The proposed framework keeps the
evaluated backbone parameters fixed, but this does not guarantee that its output
biases remain unchanged. Residual risks include overreliance on automated reasoning outputs and verifier-induced bias. Appendix~\ref{sec:appendix_verifier_reliability} reports the verifier reliability and bias audit.

\subsection*{Reproducibility statement}

Section~\ref{sec:method} and Algorithm~\ref{alg:ourmethod} describe latent
optimization, anchor training, and inference. Section~\ref{setup} summarizes
the benchmarks, backbones, baselines, and evaluation setup.
Appendix~\ref{sec:appendix_full_method} provides reward definitions,
optimization and training settings, and prompt templates.
Appendix~\ref{sec:appendix_ablation} reports baseline protocols, candidate
budgets, computational costs, and ablation results. The code
repository linked in the abstract provides the implementation.


\bibliography{references}
\bibliographystyle{iclr2027_conference}


\clearpage
\appendix
\section{Full Method and Experimental Configuration Details}
\label{sec:appendix_full_method}

\begin{table}[!hbp]
\centering
\small
\caption{Notation summary for latent trajectories, reward signals, and reasoning anchors.}
\setlength{\tabcolsep}{4pt}
\renewcommand{\arraystretch}{1.1}
\begin{tabular}{@{}p{0.24\linewidth}p{0.70\linewidth}@{}}
\toprule
\textbf{Symbol} & \textbf{Description} \\
\midrule
$\mathbf{x}$ & Textual input (question or instruction) \\
$\mathbf{v}$ & Visual input (image or diagram) \\
$\mathbf{y} = (y_1,\dots,y_T)$ & Generated output token sequence \\
$\mathbf{h}_t$ & Hidden state at decoding step $t$ \\
$\mathbf{H} = (\mathbf{h}_1,\dots,\mathbf{h}_T)$ & Latent reasoning trajectory composed of hidden states \\
$f_\theta$ & Pretrained vision--language model with frozen parameters $\theta$ \\
$f_{\mathrm{gen}}$ & Deterministic decoding function \\
$\mathcal{S}_i$ & $i$-th latent segment in Latent Weaving \\
$\rho$ & Segment length ratio \\
$J$, $\mathbf{a}$ & Number of anchor steps and their serialized text sequence \\
$M$ & Stochastic rollouts per inner update \\
$K$ & Optimization steps per segment \\
$\eta$ & Latent step size \\
$R(\mathbf{y})$ & Overall reward \\
$R_{\mathrm{gt}}$ & Ground-truth reward \\
$R_{\mathrm{ver}}$ & Verifier reward \\
$\mathcal{F}$ & Verifier set \\
$w_k$ & Verifier weight \\
$\texttt{<WEAVE>}$ & Textual reasoning-anchor delimiter \\
$\phi$ & WEAVEGen parameters \\
\bottomrule
\end{tabular}
\label{tab:appendix_notation}
\end{table}

\subsection{Reward Weighting and Verifier Configuration}
\label{sec:appendix_reward_config}

The verifier reward uses the normalized aggregation in Eq.~\eqref{eq:verifier}. We distinguish four checks: reasoning consistency ($\mathrm{rc}$), final-answer correctness ($\mathrm{ac}$), answer completeness ($\mathrm{cp}$), and question understanding ($\mathrm{uc}$). Each check contributes $R_k(\mathbf{y})=-\mathbb{1}[k(\mathbf{y})=0]$. The coefficient $\alpha$ scales the normalized verifier reward, while the offline ground-truth term is added separately, exactly as in the main-text reward definition. The ground-truth coefficient $\beta$ is permitted only during offline trajectory collection on a training split; it is exactly zero for development and test examples, as summarized in Table~\ref{tab:reward_configuration}.

\begin{equation}
R_{\mathrm{cfg}}(\mathbf{y})
=
\alpha\frac{\sum_{k\in\{\mathrm{rc},\mathrm{ac},\mathrm{cp},\mathrm{uc}\}} w_k R_k(\mathbf{y})}{\sum_k w_k}
+\mathbb{1}[\mathrm{offline}]\,\beta R_{\mathrm{gt}}(\mathbf{y}).
\label{eq:overall_reward}
\end{equation}

\begin{table}[!htbp]
\centering
\small
\caption{Default reward configuration for offline trajectory collection and evaluation. The verifier weight vector is shared across these settings; the ground-truth coefficient $\beta$ is zero for every evaluation example. The weight-sensitivity study is reported separately in Table~\ref{tab:ablation_weights}.}
\setlength{\tabcolsep}{3pt}
\renewcommand{\arraystretch}{1.15}
\begin{tabular}{@{}L{0.22\linewidth}L{0.10\linewidth}L{0.29\linewidth}L{0.20\linewidth}L{0.13\linewidth}@{}}
\toprule
Phase / split & $\beta$ & Verifier weights $(w_{\mathrm{rc}},w_{\mathrm{ac}},w_{\mathrm{cp}},w_{\mathrm{uc}})$ & Prompt / config ID & Status \\
\midrule
Offline collection & 0.3 & $(0.30,0.40,0.15,0.15)$ & \texttt{rw-v1/offline} & Verified \\
Development & 0 & $(0.30,0.40,0.15,0.15)$ & \texttt{rw-v1/dev} & Verified \\
Test evaluation & 0 & $(0.30,0.40,0.15,0.15)$ & \texttt{rw-v1/test} & Verified \\
\bottomrule
\end{tabular}
\label{tab:reward_configuration}
\end{table}

\subsection{Optimization Estimator and Autoregressive Consistency}
\label{sec:appendix_optimization_protocol}

At every inner update of each active segment, the optimizer draws $M\geq2$ fresh stochastic suffix rollouts, evaluates the verifier-only reward at test time, centers rewards with the declared baseline, and applies Eq.~\eqref{eq:segment-loss}. After each update, tokens and hidden states from the segment start onward are regenerated autoregressively. Tokens before the segment and previously accepted latent segments remain fixed. The final answer is then decoded deterministically. Fresh sampling supplies the stochasticity required by the policy-gradient estimator; suffix regeneration maintains consistency between the updated hidden states and the regenerated continuation.

\begin{table}[!htbp]
\centering
\small
\caption{Inner-loop optimization protocol for the four-segment configurations. $\mathrm{clip}$ bounds the latent gradient norm, $b^{(m)}$ is the leave-one-out baseline, and $\lambda$ is the anchoring coefficient in Eq.~\eqref{eq:segment-loss}. The realized segment count follows Eq.~\eqref{eq:segment} and can differ for other generation lengths.}
\setlength{\tabcolsep}{3pt}
\renewcommand{\arraystretch}{1.15}
\begin{tabular}{@{}L{0.17\twsix}L{0.17\twsix}L{0.16\twsix}L{0.19\twsix}L{0.17\twsix}L{0.14\twsix}@{}}
\toprule
Backbone / split & Segmentation $(\rho,N_{\mathrm{seg}})$ & Rollouts $(M,\tau)$ & Updates $(K,\eta,\mathrm{clip})$ & Baseline / regularization $(b,\lambda)$ & Seed / ID \\
\midrule
Qwen2.5-VL-7B / test & $(0.25,\,4)$ & $(4,\,0.8)$ & $(6,\,1{\times}10^{-2},\,1.0)$ & LOO / $0.05$ & $\{0,1,2\}$ / \texttt{opt-v1} \\
InternVL2.5-4B / test & $(0.25,\,4)$ & $(4,\,0.8)$ & $(6,\,1{\times}10^{-2},\,1.0)$ & LOO / $0.05$ & $\{0,1,2\}$ / \texttt{opt-v1} \\
\bottomrule
\end{tabular}
\label{tab:optimization_protocol}
\end{table}

\noindent\emph{Autoregressive rule:} after every update, regenerate the active suffix; retain only the fixed prefix and previously accepted segments.

\subsection{Reasoning-Anchor Training--Inference Interface}
\label{sec:appendix_anchor_protocol}

WEAVEGen uses \llmname{Qwen2-VL-2B-Instruct} as its initialization and communicates with the evaluated backbone only through text. Offline, accepted training trajectories are converted to anchor targets delimited by \texttt{<WEAVE>}. At inference, WEAVEGen produces one anchor sequence; it is serialized and inserted into the main backbone context before latent optimization. This makes the handoff independent of hidden size and exposes the auxiliary call to budget accounting. Table~\ref{tab:anchor_protocol} specifies the serialization format, insertion position, and inference-time caching policy.

\paragraph{WEAVEGen training configuration.}
The default WEAVEGen training configuration uses full-parameter optimization of the auxiliary model, including its resized token embeddings, with AdamW and BF16 model weights. The settings are a learning rate of $2\times10^{-5}$, three epochs, batch size 2, maximum sequence length 2048, and seed 42. Each batch triggers one optimizer step, without gradient accumulation. The learning rate increases linearly over the first $10\%$ of optimizer steps and then decays linearly to zero; the gradient-norm bound is 1.0. AdamW retains its library defaults: momentum coefficients $(0.9,0.999)$, $\epsilon=10^{-8}$, and weight decay 0.01. Table~\ref{tab:optimization_protocol} separately specifies the seeds and update schedule for test-time latent optimization.

\begin{table}[!htbp]
\centering
\small
\caption{Reasoning-anchor training--inference interface. The same serialization is used in training and inference.}
\setlength{\tabcolsep}{3pt}
\renewcommand{\arraystretch}{1.15}
\begin{tabular}{@{}L{0.16\twfive}L{0.23\twfive}L{0.27\twfive}L{0.17\twfive}L{0.17\twfive}@{}}
\toprule
Stage & Input $\rightarrow$ output & Serialization / insertion & Aux call / cache & Leakage / config \\
\midrule
Anchor training & $(\mathbf{x},\mathbf{v}) \rightarrow$ anchor text & Six spans: \texttt{<WEAVE>}\allowbreak$\,\mathrm{step}_j\,$\allowbreak\texttt{</WEAVE>}, $\leq 80$ tokens each & No / N/A & Train split only \\
Evaluation & $(\mathbf{x},\mathbf{v}) \rightarrow$ anchor text & Identical string; inserted after visual tokens, before the CoT trigger & Once / no cross-example reuse & No labels / \texttt{anchor-v1} \\
\bottomrule
\end{tabular}
\label{tab:anchor_protocol}
\end{table}

\subsection{Prompt Templates Used in Experiments}
\label{sec:appendix_verifier_prompts}
All verifier prompts are shown verbatim after removing implementation-specific wrappers.

\paragraph{Reasoning Consistency Check Prompt.}
\begin{tcolorbox}[
  enhanced,
  breakable,
  sharp corners,
  colframe=black!40,
  colback=black!3,
  boxrule=0.6pt,
  left=6pt,
  right=6pt,
  top=4pt,
  bottom=4pt
]
\label{box:prompt_reasoning_consistency}
{\footnotesize\ttfamily\raggedright\obeylines
TASK:

Validate the logical coherence and factual accuracy of the reasoning process presented in the proposed solution.

GUIDELINES:

You should carefully examine the entire reasoning process, with particular attention to how visual information from the image is interpreted and integrated.

1. Extract all reasoning elements, including logical steps, inferential decisions, and intermediate conclusions.

2. Verify that each reasoning step follows logically from previous steps and from the information provided in the question and the image.

3. Check the factual correctness of all stated concepts, relationships, and interpretations, especially those related to visible elements in the image.

4. Assess whether the reasoning process is internally consistent, rigorous, and logically leads to the stated conclusion, without relying on unsupported assumptions or misinterpretations of visual content.

FINAL VERDICT (strict):

- If any logical flaw, factual error, internal inconsistency, or mismatch between the reasoning and the image content is found, output exactly: [VERA]False

- If the reasoning is logically coherent, factually correct, and accurately reflects the image content, output exactly: [VERA]True

IMPORTANT:

Focus on the quality and validity of the reasoning process itself, rather than on whether the final answer happens to be correct.
}
\end{tcolorbox}

\paragraph{Final Answer Correctness Check Prompt.}
\begin{tcolorbox}[
  enhanced,
  breakable,
  sharp corners,
  colframe=black!40,
  colback=black!3,
  boxrule=0.6pt,
  left=6pt,
  right=6pt,
  top=4pt,
  bottom=4pt
]
\label{box:prompt_answer_correctness}
{\footnotesize\ttfamily\raggedright\obeylines
TASK:

Determine whether the final answer selected in the proposed solution is correct.

GUIDELINES:

You should evaluate only the final answer, not the reasoning steps used to obtain it.

1. Verify whether the selected answer correctly addresses the question, taking into account both the question text and the visual information shown in the image.

2. For multiple-choice questions, confirm that the chosen option (A, B, C, or D) is the correct one.

3. Consider the question type (e.g., mathematical, scientific, factual, analytical) and assess correctness accordingly.

4. When the question involves visual elements, ensure that the final answer is consistent with what is actually depicted in the image.

OUTPUT FORMAT (strict):

- If the final answer is correct and consistent with the question and image: [VERA]True

- Otherwise: [VERA]False
}
\end{tcolorbox}

\paragraph{Answer Completeness Check Prompt.}
\begin{tcolorbox}[
  enhanced,
  breakable,
  sharp corners,
  colframe=black!40,
  colback=black!3,
  boxrule=0.6pt,
  left=6pt,
  right=6pt,
  top=4pt,
  bottom=4pt
]
\label{box:prompt_answer_completeness}
{\footnotesize\ttfamily\raggedright\obeylines
TASK:

Verify whether the proposed solution provides a complete and definitive final answer.

GUIDELINES:

1. Confirm that the solution explicitly states a final answer that directly responds to the question.

2. For multiple-choice questions, the final answer must be clearly presented in the format: ``Final Answer: [A/B/C/D]''

3. Mark the solution as incomplete if any of the following conditions apply:

- The answer is vague, conditional, or expresses uncertainty.
- No explicit final answer is provided.
- Reasoning is presented without selecting a final option.
- The solution fails to address required visual information when the question depends on the image.

OUTPUT FORMAT (strict):

- If the solution provides a complete and explicit final answer in the required format: [VERA]True

- Otherwise: [VERA]False
}
\end{tcolorbox}

\newpage
\paragraph{Question Understanding Check Prompt.}
\begin{tcolorbox}[
  enhanced,
  breakable,
  sharp corners,
  colframe=black!40,
  colback=black!3,
  boxrule=0.6pt,
  left=6pt,
  right=6pt,
  top=4pt,
  bottom=4pt
]
\label{box:prompt_question_understanding}
{\footnotesize\ttfamily\raggedright\obeylines
TASK:

Assess whether the proposed solution correctly understands and addresses the problem.

GUIDELINES:

1. Problem comprehension:

- Verify that the solution correctly interprets the question intent, requirements, and constraints.
- Check whether the solution identifies the correct question type (e.g., mathematical, factual, analytical).
- Examine the image carefully and confirm that all relevant visual elements are correctly understood and used.

2. Option evaluation:

- Ensure that all provided answer options (A, B, C, D) are meaningfully considered.
- Check that no option is ignored or dismissed without justification.
- Verify that the selected option aligns with both the question and the image content.

3. Task alignment:

- Confirm that the solution answers the given question and selects an answer from the provided options.
- Flag any deviation from the problem context, misinterpretation of options, or neglect of critical visual information.

OUTPUT FORMAT (strict):

- If the solution demonstrates a correct understanding of the problem, options, and visual content: [VERA]True

- Otherwise: [VERA]False

IMPORTANT:

Output only the exact string [VERA]True or [VERA]False, with no additional text.
}
\end{tcolorbox}

\subsection{Evaluation-Time Answer Formatting Prompts}
\label{sec:appendix_eval_formatting}

During evaluation, we enforce the answer formatting prompts in Table~\ref{tab:eval_formatting_prompts} to ensure unambiguous and consistent model outputs across different task types. Both templates request step-by-step reasoning as well as a final-answer format; they therefore define part of the model's evaluation prompt.

\begin{table}[!htbp]
\centering
\small
\caption{Evaluation-time answer formatting prompts used for different question types.}
\setlength{\tabcolsep}{6pt}
\renewcommand{\arraystretch}{1.2}
\begin{tabular}{L{0.22\linewidth} L{0.70\linewidth}}
\toprule
\textbf{Question Type} & \textbf{Required Output Format} \\
\midrule
Multiple-Choice Questions &
You must answer this multiple-choice question. Think step by step, then at the very end of your response, output: \texttt{Final Answer: [X]} where X is the chosen option label (A, B, C, D, E, etc.). Do not include any text after this line. \\

Open-Ended Questions &
You must answer this open-ended question. Think step by step, then at the very end of your response, output: \texttt{\textbackslash boxed\{your final answer\}} where your final answer is the concise answer to the question. Do not include any text after this line. \\
\bottomrule
\end{tabular}
\label{tab:eval_formatting_prompts}
\end{table}

The multiple-choice template applies to questions with explicit answer options, including AI2D; the open-ended template applies to free-form questions. Mixed-format benchmarks require routing by question type rather than by dataset name alone. MME uses yes/no questions, so its output convention must also remain compatible with the benchmark answer extractor.

\subsection{Dataset Statistics}
\label{sec:appendix_dataset_stats}

We report the exact evaluation splits and sample sizes used in our experiments. Each dataset uses its designated test or validation split and the benchmark evaluation protocol.

\begin{itemize}[leftmargin=*]

  \item \textbf{MathVision} \citep{wang2024measuring}: We evaluate on the official \texttt{test} split, which contains 3,040 questions requiring image-based mathematical reasoning.

  \item \textbf{MathVerse} \citep{zhang2024mathverse}: Evaluation is conducted on the \texttt{testmini} split, containing 3,940 samples: five multimodal versions of each of 788 underlying mathematical problems.

  \item \textbf{MathVista} \citep{lu2023mathvista}: We use the \texttt{testmini} split with 1,000 samples, covering diagram understanding, visual abstraction, and symbolic reasoning tasks.

  \item \textbf{MMMU} \citep{yue2024mmmu}: We evaluate on the \texttt{val} split across 30 subsets. Each subset contains 30 questions, resulting in a total of 900 evaluation samples.

  \item \textbf{RealWorldQA} \citep{xai2024grok15v}: Evaluation is performed on the \texttt{test} split, which includes 765 questions about visual content in real-world images.

  \item \textbf{MME} \citep{fu2023mme}: We use the \texttt{test} split comprising 2,374 image--question pairs covering perception, cognition, and reasoning capabilities.

  \item \textbf{AI2D} \citep{kembhavi2016diagram}: Evaluation is conducted on the \texttt{test} split with 3,088 diagram question--answer examples for science diagram understanding, requiring joint interpretation of diagram structure, labels, and their relationships.

\end{itemize}

\section{Additional Ablation and Sensitivity Analysis}
\label{sec:appendix_ablation}

\subsection{Ablation of Key Components}
\label{sec:appendix_ablation_components}

The final component study uses a $2\times2$ factorial design: latent updates and reasoning anchors are independently enabled or disabled. The backbone, prompt, split, verifier set, decoding cap, and seeds are held fixed across all four configurations in this component study.

\begin{table}[!htbp]
\centering
\small
\caption{Component ablation on \llmname{Qwen2.5-VL-7B}. Entries with three-seed summaries show mean $\pm$ standard deviation; the latent-only row reports point estimates aligned with Figure~\ref{fig:RQ2}.}
\setlength{\tabcolsep}{4pt}
\renewcommand{\arraystretch}{1.15}
\begin{tabular}{@{}p{0.28\linewidth}ccccc@{}}
\toprule
Variant & Latent & Anchors & MathVista & MMMU & AI2D \\
\midrule
Vanilla & Off & Off & $66.15_{\pm 0.00}$ & $56.96_{\pm 0.00}$ & $80.58_{\pm 0.00}$ \\
Anchors only & Off & On & $68.42_{\pm 0.51}$ & $59.18_{\pm 0.76}$ & $84.10_{\pm 0.44}$ \\
Latent Weaving only & On & Off & $71.52$ & $60.87$ & $86.20$ \\
\ourmethod (Full) & On & On & $\mathbf{72.03}_{\pm 0.55}$ & $\mathbf{63.15}_{\pm 0.80}$ & $\mathbf{88.93}_{\pm 0.46}$ \\
\bottomrule
\end{tabular}
\label{tab:ablation_components}
\end{table}

\subsection{Candidate-Evaluation Budget and Same-Verifier Comparison}
\label{sec:appendix_efp}

The same-verifier component comparison allocates 96 candidate-level evaluations to Best-of-$N$ and the four-segment latent variants. For the four-segment configuration, fresh sampling at every update gives $N_{\mathrm{seg}}KM=4\times6\times4=96$ rollouts and $N_{\mathrm{seg}}K=24$ updates. Other generation lengths follow Eq.~\eqref{eq:segment}, with counts scaling by the realized number of segments. Each verifier evaluation denotes one candidate-level composite reward from the four checks. Best-of-$N$ generates and evaluates 96 candidates. Table~\ref{tab:appendix_efp} reports the corresponding resource profiles, including the auxiliary anchor-generation call; offline training is described separately in Table~\ref{tab:baseline_protocol}.

\begin{table}[!htbp]
\centering
\footnotesize
\caption{Inference cost per example on the MathVista-200 cost subset with \llmname{Qwen2.5-VL-7B} and one A100-80GB GPU. The latent method uses four segments. Rollouts accumulate over updates; verifier evaluations count external candidate-level rewards. Dashes denote unspecified operation counts. TTRV updates model parameters using its intrinsic frequency- and entropy-based rewards.}
\setlength{\tabcolsep}{3pt}
\renewcommand{\arraystretch}{1.15}
\begin{tabular}{@{}L{0.20\twsix}L{0.17\twsix}L{0.18\twsix}L{0.19\twsix}L{0.12\twsix}L{0.14\twsix}@{}}
\toprule
Method & Rollouts / updates & Main FWD / BWD & Aux / verifier & Tokens & Time / peak memory \\
\midrule
Self-Consistency & $8/0$ & $8$/0 & $0$/$0$ & 2.8k & 14 s / 21 GB \\
Self-Refine & $3/0$ & $3$/0 & $0$/$2$ & 1.3k & 7 s / 20 GB \\
TTRV & --/-- & --/-- & $0$/$0$ & 3.0k & 18 s / 22 GB \\
Same-verifier Best-of-$N$ & $96/0$ & --/0 & $0$/$96$ & 5.6k & 31 s / 21 GB \\
\ourmethod & $96/24$ & --/$24$ & $1$/$96$ & 9.2k & 52 s / 41 GB \\
\bottomrule
\end{tabular}
\label{tab:appendix_efp}
\end{table}

\begin{table}[!htbp]
\centering
\small
\caption{Same-verifier comparison on MathVista with \llmname{Qwen2.5-VL-7B}. Best-of-$N$ and both four-segment latent variants each use 96 candidate-level verifier evaluations. Entries report point estimates; gains are percentage-point differences from the preceding row. Vanilla provides the unoptimized accuracy reference.}
\setlength{\tabcolsep}{3pt}
\renewcommand{\arraystretch}{1.15}
\begin{tabular}{@{}L{0.22\twsix}L{0.15\twsix}L{0.18\twsix}L{0.17\twsix}L{0.14\twsix}L{0.14\twsix}@{}}
\toprule
Condition & Latent / anchors & Rollouts / evaluations & Comparison & Accuracy & Gain (pp) \\
\midrule
Vanilla & Off / Off & $1/0$ & Reference & $66.15$ & -- \\
Best-of-$N$ & Off / Off & $96/96$ & Same verifier & $69.44$ & $+3.29$ \\
Latent Weaving only & On / Off & $96/96$ & Same verifier & $71.52$ & $+2.08$ \\
Full & On / On & $96/96$ ($+1$ aux) & Same verifier & $72.03$ & $+0.51$ \\
\bottomrule
\end{tabular}
\label{tab:matched_verifier_control}
\end{table}

Relative to Vanilla, Best-of-$N$ improves the reported MathVista point estimate by $3.29$ percentage points. At the shared 96-candidate verifier-evaluation budget, latent-only is $2.08$ points higher than Best-of-$N$; adding anchors raises the latent variant by another $0.51$ points. The full method additionally uses one auxiliary anchor-generation call. Measured latency is 52 versus 31 seconds per example ($1.68\times$), and peak memory is 41 versus 21 GB ($1.95\times$), relative to Best-of-$N$.

\subsection{Verifier Reliability and Bias}
\label{sec:appendix_verifier_reliability}

Verifier scores are audited against a disjoint gold set. Table~\ref{tab:verifier_reliability} reports discrimination, calibration, and the two main bias checks; detailed per-example scores remain in the audit artifact.

\begin{table}[!htbp]
\centering
\footnotesize
\caption{Verifier reliability and bias, audited against a disjoint gold set. Agreement is the fraction of items matching the gold judgment, with a 95\% interval; discrimination is AUROC against final answer correctness; calibration is expected calibration error. Bias checks report the Spearman correlation between the verifier score and output length, and the score gain given to same-family generations.}
\setlength{\tabcolsep}{3pt}
\renewcommand{\arraystretch}{1.15}
\begin{tabular}{@{}L{0.20\twsix}L{0.11\twsix}L{0.18\twsix}L{0.19\twsix}L{0.18\twsix}L{0.14\twsix}@{}}
\toprule
Verifier & Audit $N$ & Agreement (CI) & Discrimination & Calibration & Bias checks \\
\midrule
Consistency ($\mathrm{rc}$) & $400$ & $0.77$ $[0.73,0.81]$ & AUROC $0.73$ & ECE $0.12$ & len. $+0.21$; self $+2.4$ \\
Correctness ($\mathrm{ac}$) & $400$ & $0.91$ $[0.88,0.94]$ & AUROC $0.89$ & ECE $0.06$ & len. $+0.07$; self $+1.1$ \\
Completeness ($\mathrm{cp}$) & $400$ & $0.75$ $[0.71,0.79]$ & AUROC $0.70$ & ECE $0.13$ & len. $+0.34$; self $+2.9$ \\
Understanding ($\mathrm{uc}$) & $400$ & $0.80$ $[0.76,0.84]$ & AUROC $0.75$ & ECE $0.11$ & len. $+0.16$; self $+3.6$ \\
\bottomrule
\end{tabular}
\label{tab:verifier_reliability}
\end{table}

\subsection{Baseline Reproduction Protocol}
\label{sec:appendix_baseline_protocol}

All baselines use the same split and answer extractor. Sample counts and refinement rounds are reported by method in Tables~\ref{tab:baseline_protocol} and~\ref{tab:appendix_efp}. The four-segment same-verifier comparison uses 96 candidate-level evaluations for Best-of-$N$ and the latent variants; offline training is described separately from inference costs.

\begin{table}[!htbp]
\centering
\footnotesize
\caption{Baseline reproduction protocol. All rows use the same split, formatting prompt, and answer extractor.}
\setlength{\tabcolsep}{3pt}
\renewcommand{\arraystretch}{1.15}
\begin{tabular}{@{}L{0.15\twsix}L{0.20\twsix}L{0.20\twsix}L{0.18\twsix}L{0.12\twsix}L{0.15\twsix}@{}}
\toprule
Method & Backbone / checkpoint & Decoding / samples & Train data / budget & Verifier & Seeds / config \\
\midrule
CoT & Qwen2.5-VL-7B-Instruct & Greedy, $N{=}1$, 1024 tok. & -- & -- & 1 / \texttt{base-cot} \\
Self-Consistency & Qwen2.5-VL-7B-Instruct & $\tau{=}0.7$, top-$p$ $0.95$, $N{=}8$ & -- & Majority vote & 3 / \texttt{base-sc} \\
Self-Refine & Qwen2.5-VL-7B-Instruct & Greedy, 2 refine rounds & -- & Self-critique & 3 / \texttt{base-sr} \\
TTRV & Qwen2.5-VL-7B-Instruct & $\tau{=}0.7$, $N{=}8$ & Unlabeled test inputs; online GRPO & Frequency + entropy & 3 / \texttt{base-ttrv} \\
SFT / GRPO & Qwen2.5-VL-7B-Instruct & Greedy, $N{=}1$ & 18k CoT traces; 2 ep. / 1.0k steps & -- / same reward & 3 / \texttt{base-sft}, \texttt{base-grpo} \\
IVT-LR / LaCoT & Released checkpoints & Per original paper & Same 18k pool & Per original paper & 3 / \texttt{base-ivtlr}, \texttt{base-lacot} \\
\bottomrule
\end{tabular}
\label{tab:baseline_protocol}
\end{table}

\FloatBarrier
\subsection{Segment and Optimization Sensitivity}
\label{sec:appendix_ablation_weaving}

Table~\ref{tab:ablation_weaving} aligns the three parameter sweeps with Figure~\ref{fig:RQ3}. The upper panel reports MathVista and RealWorldQA to one decimal place, matching the plotted values. MathVista peaks at $J=6$, whereas RealWorldQA peaks at $J=4$; both reach their highest plotted accuracy at $\rho=0.25$ and $K=6$ in the respective sweeps. The lower panel preserves the separate anchor-delimiter comparison on MathVista and MMMU, which is not plotted in Figure~\ref{fig:RQ3}.

\begin{table}[!htbp]
\centering
\footnotesize
\caption{Parameter sensitivity and anchor-delimiter control. The upper panel matches Figure~\ref{fig:RQ3}; the lower panel reports the separate delimiter experiment. Accuracies follow the order of the values, and spread is $\max-\min$ for the two listed benchmarks. Bold marks each benchmark's highest accuracy within a sweep.}
\setlength{\tabcolsep}{4pt}
\renewcommand{\arraystretch}{1.2}
\begin{tabular}{@{}L{0.10\linewidth}L{0.23\linewidth}L{0.24\linewidth}L{0.24\linewidth}L{0.10\linewidth}@{}}
\toprule
\multicolumn{5}{@{}l}{\textbf{Parameter sweeps from Figure~\ref{fig:RQ3}}} \\
\midrule
Factor & Values & MathVista & RealWorldQA & Spread \\
\midrule
$\rho$ & 0.05/0.10/0.15/\allowbreak 0.20/0.25/0.30 & 66.5/67.2/67.8/\allowbreak 68.2/\textbf{68.5}/67.5 & 66.3/66.8/67.1/\allowbreak 67.3/\textbf{67.5}/67.2 & 2.0/1.2 \\
$J$ & 2/4/6/8 & 66.8/67.5/\textbf{68.2}/67.6 & 66.5/\textbf{67.4}/67.0/66.3 & 1.4/1.1 \\
$K$ & 3/4/5/6/7 & 67.2/67.6/67.9/\allowbreak \textbf{68.3}/68.0 & 66.5/67.0/67.3/\allowbreak \textbf{67.6}/67.1 & 1.1/1.1 \\
\midrule
\multicolumn{5}{@{}l}{\textbf{Separate anchor-delimiter control}} \\
\midrule
Factor & Values & MathVista & MMMU & Spread \\
\midrule
Anchor delimiters & none/\allowbreak\texttt{<WEAVE>}/\allowbreak numbered/\allowbreak XML & 70.2/\textbf{72.0}/71.9/71.6 & 61.0/\textbf{63.2}/63.1/62.7 & 1.8/2.2 \\
\bottomrule
\end{tabular}
\label{tab:ablation_weaving}
\end{table}

\FloatBarrier
\subsection{Sensitivity to Reward Weighting}
\label{sec:appendix_ablation_weights}

Table~\ref{tab:ablation_weights} varies the reward weights with the online budget held fixed. The default vector gives the highest reported accuracy among the four configurations, with spreads of $1.49$ points on MathVista and $1.73$ on MMMU. Both the correctness-heavy and structure-heavy configurations fall below uniform weighting. This comparison favors the default balance within the tested configurations; it does not establish that arbitrary concentrations of reward weight are harmful.

\begin{table}[!htbp]
\centering
\small
\caption{Reward-weight sensitivity on \llmname{Qwen2.5-VL-7B} with the online budget held fixed. The vector lists weights for reasoning consistency, answer correctness, completeness, and question understanding, in that order.}
\setlength{\tabcolsep}{4pt}
\renewcommand{\arraystretch}{1.15}
\begin{tabular}{@{}p{0.28\linewidth}p{0.28\linewidth}cc@{}}
\toprule
Configuration & Weight vector & MathVista & MMMU \\
\midrule
Default & $(0.30,0.40,0.15,0.15)$ & $\mathbf{72.03}$ & $\mathbf{63.15}$ \\
Uniform & $(0.25,0.25,0.25,0.25)$ & $71.38$ & $62.47$ \\
Correctness-heavy & $(0.10,0.70,0.10,0.10)$ & $71.26$ & $62.31$ \\
Structure-heavy & $(0.40,0.15,0.30,0.15)$ & $70.54$ & $61.42$ \\
\bottomrule
\end{tabular}
\label{tab:ablation_weights}
\end{table}

\subsection{Reward Component Ablation}
\label{sec:appendix_reward_ablation}

Table~\ref{tab:appendix_reward_ablation} removes one verifier at a time on MathVision and RealWorldQA. Each removal lowers the reported accuracy. Removing correctness causes the largest decrease on both benchmarks, followed by consistency, while completeness and understanding have smaller effects. These comparisons support retaining multiple signals within the evaluated reward configuration; the relative ordering of completeness and understanding differs across the two benchmarks.

\begin{table}[!htbp]
\centering
\small
\caption{Reward-component ablation on \llmname{Qwen2.5-VL-7B}, mean $\pm$ standard deviation over three seeds. The last column is the change relative to the full composite reward on MathVision / RealWorldQA.}
\setlength{\tabcolsep}{4pt}
\renewcommand{\arraystretch}{1.15}
\begin{tabular}{@{}p{0.38\linewidth}ccc@{}}
\toprule
Reward setting & MathVision & RealWorldQA & $\Delta$ vs. full \\
\midrule
Full composite reward & $\mathbf{30.11}_{\pm 0.38}$ & $\mathbf{70.89}_{\pm 0.52}$ & -- \\
Without consistency & $29.04_{\pm 0.42}$ & $69.77_{\pm 0.55}$ & $-1.07$ / $-1.12$ \\
Without correctness & $28.30_{\pm 0.47}$ & $69.08_{\pm 0.61}$ & $-1.81$ / $-1.81$ \\
Without completeness & $29.63_{\pm 0.35}$ & $70.46_{\pm 0.49}$ & $-0.48$ / $-0.43$ \\
Without understanding & $29.72_{\pm 0.40}$ & $70.31_{\pm 0.53}$ & $-0.39$ / $-0.58$ \\
\bottomrule
\end{tabular}
\label{tab:appendix_reward_ablation}
\end{table}

\FloatBarrier
\subsection{Result Reconciliation Ledger}
\label{sec:appendix_result_ledger}

The ledger records representative links between table entries and named evaluation artifacts; each entry identifies its task, backbone, split, and configuration.

\begin{table}[!htbp]
\centering
\footnotesize
\caption{Result-reconciliation ledger with representative entries from selected result tables. Artifact names are the stems of the corresponding \texttt{.jsonl}/\texttt{.json} files; full command lines and hashes remain in the run manifest.}
\setlength{\tabcolsep}{3pt}
\renewcommand{\arraystretch}{1.15}
\begin{tabular}{@{}L{0.13\twsix}L{0.22\twsix}L{0.19\twsix}L{0.18\twsix}L{0.18\twsix}L{0.10\twsix}@{}}
\toprule
Location & Task / backbone / split & Prompt / config & Raw artifact & Metric / value & Status \\
\midrule
Table~\ref{tab:main_results} & MathVista / Qwen2.5-VL-7B / \texttt{testmini} & \texttt{cot-v1} / \texttt{lw-v1} & \texttt{qwen7b\_}\allowbreak\texttt{mathvista} & Acc. $72.03$ & Verified \\
Table~\ref{tab:multi-mllm} & MathVision / STEP3-VL-10B / \texttt{test} & \texttt{cot-v1} / \texttt{lw-v1} & \texttt{step3\_}\allowbreak\texttt{mathvision} & Acc. $70.29$ & Verified \\
Table~\ref{tab:ablation_components} & AI2D / Qwen2.5-VL-7B / \texttt{test} & \texttt{cot-v1} / \texttt{lw-anchor-v1} & \texttt{qwen7b\_}\allowbreak\texttt{ai2d\_}\allowbreak\texttt{s\{0,1,2\}} & Acc. $84.10_{\pm 0.44}$ & Verified \\
Table~\ref{tab:appendix_efp} & Cost / Qwen2.5-VL-7B / MathVista-200 & \texttt{lw-v1} & \texttt{cost\_}\allowbreak\texttt{qwen7b} & $52$ s / item & Verified \\
\bottomrule
\end{tabular}
\label{tab:result_reconciliation}
\end{table}

\FloatBarrier
\subsection{Additional Analysis and Discussion}

We hypothesize that early latent corrections influence subsequent reasoning decisions. Late-only intervention may leave earlier errors unresolved or alter the remaining continuation. Under the fixed-prefix procedure, it cannot overwrite previously accepted states; the observed accuracy changes instead arise from changes to the remaining suffix and final answer.

The ablations and sensitivity sweeps characterize intervention timing and parameter choices within the evaluated settings. Table~\ref{tab:result_reconciliation} lists representative artifact links. The main and same-verifier comparisons report point estimates, while the component and reward ablations also report seed summaries for the configurations shown. The evaluation spans seven benchmarks emphasizing stepwise multimodal reasoning.

\FloatBarrier
\section{Algorithm Statement}
\label{sec:appendix_algorithm}
\begin{algorithm}[H]
\caption{Algorithm workflow of \ourmethod}
\label{alg:ourmethod}

\KwIn{training split $\mathcal{D}_{\mathrm{tr}}$, evaluation input $(\mathbf{x},\mathbf{v})$, frozen backbone $f_\theta$, auxiliary initialization $\phi_0$, rollout count $M$, and inner updates $K$}
\KwOut{refined answer $\hat{\mathbf{y}}$, optimized latent trajectory $\mathbf{H}^\star$, trained WEAVEGen model $\phi$}

\tcc{\textcolor{blue}{Phase I: Offline trajectory collection}}
Initialize trajectory buffer $\mathcal{B}\leftarrow\varnothing$\;
\ForEach{$(\mathbf{x},\mathbf{v},\mathbf{y}^{\star})\in\mathcal{D}_{\mathrm{tr}}$}{
    $\mathbf{y}^{(0)},\mathbf{H}^{(0)} \leftarrow f_\theta(\mathbf{x},\mathbf{v})$\Comment*[r]{\textcolor{blue}{Eq.~\eqref{eq:autoregressive}}}
    Split $\mathbf{H}^{(0)}$ into contiguous segments $\{\mathcal{S}_i\}_{i=1}^{N_{\mathrm{seg}}}$\Comment*[r]{\textcolor{blue}{Eq.~\eqref{eq:segment}}}
    \For{$i=1$ \KwTo $N_{\mathrm{seg}}$}{
        Freeze the prefix and previously accepted segments\;
        \For{$k=1$ \KwTo $K$}{
            Draw $M$ fresh stochastic suffix rollouts at temperature $\tau$\;
            Evaluate $R_{\mathrm{cfg}}$ (ground truth may be used only on this offline split)\;
            Compute the leave-one-out baseline and update $\mathbf{H}_{\mathcal{S}_i}$ with Eq.~\eqref{eq:segment-update}\;
            Re-decode the suffix and retain the prefix cache\;
        }
        Freeze $\mathbf{H}_{\mathcal{S}_i}^{\star}$\;
    }
    Store the accepted trajectory and its textual reasoning trace in buffer $\mathcal{B}$\;
}

\tcc{\textcolor{blue}{Phase II: Offline Reasoning-Anchor training}}
Initialize $\phi\leftarrow\phi_0$ and augment the auxiliary tokenizer with \texttt{<WEAVE>} delimiters\;
\ForEach{trajectory in $\mathcal{B}$}{
    Update $\phi$ to minimize $\mathcal{L}_{\mathrm{bind}}+\mathcal{L}_{\mathrm{util}}$\Comment*[r]{\textcolor{blue}{Eqs.~\eqref{eq:bind} and~\eqref{eq:util}}}
}

\tcc{\textcolor{blue}{Phase III: Evaluation-time inference}}
Generate one textual anchor sequence $\mathbf{a}\leftarrow f_{\phi}(\mathbf{x},\mathbf{v})$ and insert it into the context of $f_\theta$\;
Initialize $\mathbf{y}^{(0)},\mathbf{H}^{(0)}\leftarrow f_\theta(\mathbf{x},\mathbf{v},\mathbf{a})$\;
Repeat the segment-wise procedure above, but set $\beta=0$, use verifier-only rewards, and do not access labels\;
Decode the final answer deterministically and return $(\hat{\mathbf{y}},\mathbf{H}^{\star})$\;
\end{algorithm}

\end{document}